\documentclass[10pt,twocolumn,letterpaper]{article}

\usepackage{cvpr}              

\usepackage[dvipsnames]{xcolor}

\definecolor{cvprblue}{rgb}{0.21,0.49,0.74}
\usepackage{times}
\usepackage{epsfig}
\usepackage{graphicx}
\usepackage{amsmath}
\usepackage{amssymb}
\usepackage{booktabs}

\usepackage[inline]{enumitem}

\usepackage{pifont}
\usepackage{caption}

\usepackage[accsupp]{axessibility}  

\newenvironment{tight_enumerate}{
\begin{enumerate}[leftmargin=15pt]
  \setlength{\topsep}{0pt}
  \setlength{\itemsep}{0pt}
  \setlength{\parskip}{0pt}
  \setlength{\parsep}{0pt}
}{\end{enumerate}}

\usepackage{subcaption}
\usepackage{booktabs}       

\usepackage{tabularx}
\usepackage{multirow}
\usepackage{xcolor}
\newcommand{\Ours}{UDOS}
\newcommand{\figref}[1]{Fig.~\ref{#1}}
\newcommand{\tabref}[1]{Tab.~\ref{#1}}
\newcommand{\secref}[1]{Sec.~\ref{#1}}

\renewcommand{\eqref}[1]{Eq.~(\ref{#1})}
\newcommand{\supp}{\textcolor{black}{supplementary}}
\newcommand{\tick}{{{{\ding{51}}}}}
\newcommand{\cross}{{{{\ding{55}}}}}
\newcommand{\noop}[1]{#1}

\newcommand{\review}[1]{#1}

\usepackage[pagebackref,breaklinks=true,bookmarks=false]{hyperref}

\usepackage[capitalise,noabbrev,nameinlink]{cleveref}

\usepackage{abstract}

\def\paperID{32} 
\def\confName{CVPR}
\def\confYear{2024}

\title{Open-world Instance Segmentation: Top-down Learning with Bottom-up Supervision}

\newcommand{\MYhref}[3][blue]{\href{#2}{\color{#1}{#3}}}%
\author{
    Tarun Kalluri\textsuperscript{$\dagger$}\thanks{Work done during TK's internship at Meta.} \qquad Weiyao Wang\textsuperscript{$\ddagger$} \qquad Heng Wang\textsuperscript{$\ddagger$} \qquad Manmohan Chandraker\textsuperscript{$\dagger$} \\
    Lorenzo Torresani\textsuperscript{$\ddagger$} \qquad Du Tran\textsuperscript{$\ddagger$} \\ \\
    \textsuperscript{$\dagger$}UC San Diego \quad 
    \textsuperscript{$\ddagger$} Meta AI \\
    \MYhref[violet]{https://tarun005.github.io/UDOS}{\texttt{https://tarun005.github.io/UDOS}}
}

\begin{document}

\twocolumn[{\renewcommand\twocolumn[1][]{#1}\maketitle
\begin{center}
    \centering
    \captionsetup{type=figure}
    \includegraphics[width=0.95\textwidth]{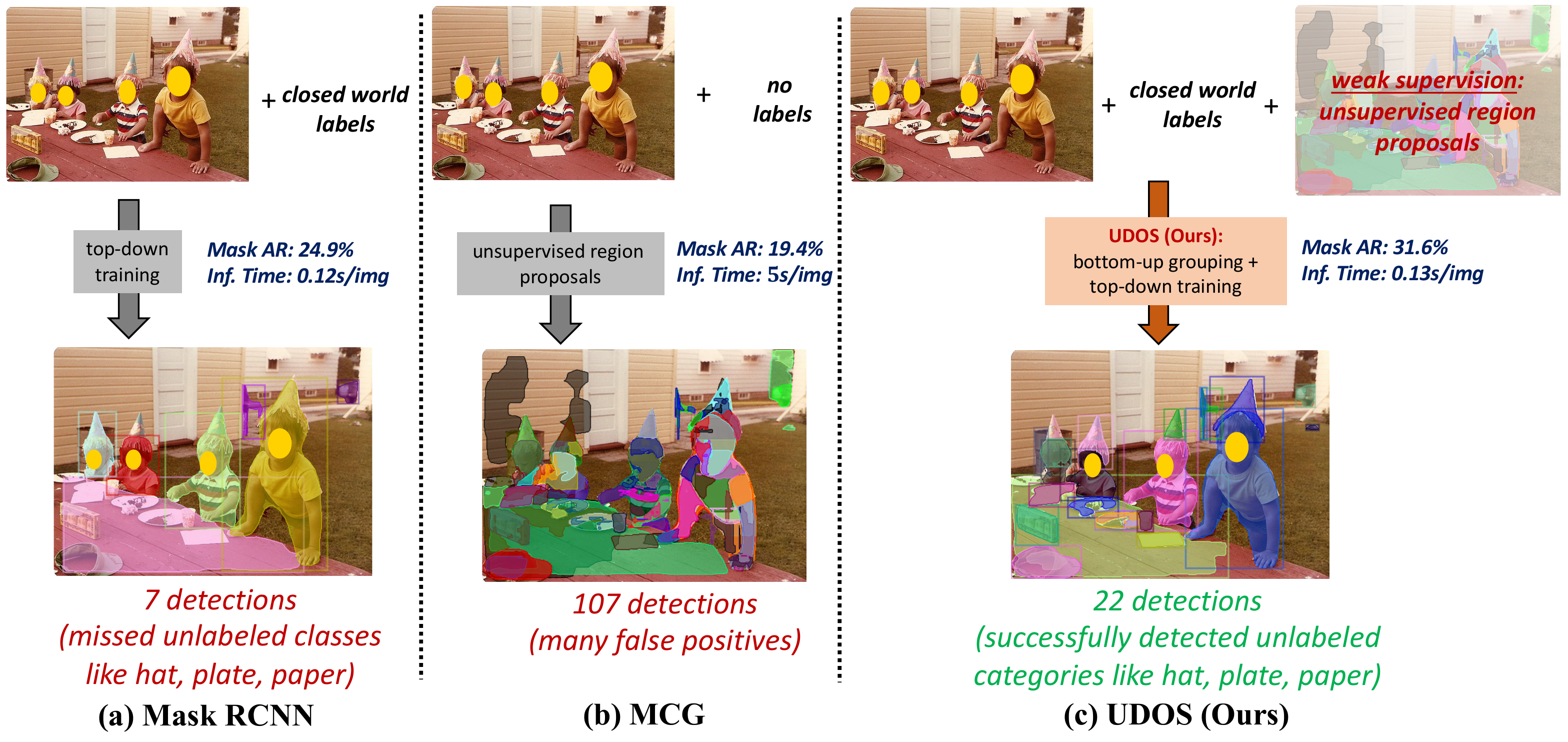}
    \captionof{figure}{{\bf Open world segmentation using \Ours{}}. Image from COCO. \textbf{(a)} Mask R-CNN~\citep{he2017mask}, trained on VOC-categories from COCO, fails to detect many unseen categories due to seen-class bias; \textbf{(b)} MCG~\citep{pont2016multiscale} provides diverse proposals, but predicts many over-segmented false-positives with noisy boundaries; \textbf{(c)} combining the advantages of (a) and (b) into a joint framework, \Ours{} efficiently detects unseen classes in open world when trained only using VOC-categories from COCO, while adding negligible inference time overhead.
        }
    \label{fig:intro}
    \vspace{-2pt}
\end{center}
}]

\saythanks
\begin{abstract}
Top-down instance segmentation architectures excel with predefined closed-world taxonomies but exhibit biases and performance degradation in open-world scenarios. In this work, we introduce bottom-\textbf{U}p and top-\textbf{D}own \textbf{O}pen-world \textbf{S}egmentation (\Ours{}), a novel approach that combines classical bottom-up segmentation methods within a top-down learning framework. \Ours{} leverages a top-down network trained with weak supervision derived from class-agnostic bottom-up segmentation to predict object parts. These part-masks undergo affinity-based grouping and refinement to generate precise instance-level segmentations. \Ours{} balances the efficiency of top-down architectures with the capacity to handle unseen categories through bottom-up supervision. We validate \Ours{} on challenging datasets (MS-COCO, LVIS, ADE20k, UVO, and OpenImages), achieving superior performance over state-of-the-art methods in cross-category and cross-dataset transfer tasks. Our code and models will be publicly available.
\end{abstract}

\section{Introduction}
\label{sec:intro}

Open world instance segmentation~\citep{wang2021unidentified} is the task of predicting class-agnostic instance masks for all objects within an image. A pivotal challenge therein lies in effectively segmenting novel instances, i.e., instances from categories not in the training taxonomy. This capability assumes paramount importance for ensuring the robust and dependable real-world deployment of instance segmentation models across domains like robotics~\citep{xie2021unseen}, autonomous driving~\citep{neuhold2017mapillary, cordts2016cityscapes}, and embodied AI~\citep{savva2019habitat}, where novel objects are encountered regularly. While expanding taxonomy during annotation is a potential countermeasure, it presents notable challenges: it necessitates substantial human effort to amass sufficient annotations for each category, and achieving a comprehensive taxonomy encompassing \textit{all} conceivable categories remains an impractical endeavor. Consequently, the emphasis remains on the model's capacity to generalize and proficiently segment novel objects—a more pragmatic approach.


Common instance segmentation frameworks like Mask R-CNN~\citep{he2017mask} often tightly couple recognition and segmentation~\citep{wang2021unidentified}, making it challenging to accurately segment objects not present in the training data. This issue is particularly pronounced when these frameworks are trained with non-exhaustive annotations like MS-COCO~\citep{lin2014microsoft}, where out-of-taxonomy objects are treated as background, resulting in penalties for predictions made on these objects. In \figref{fig:intro}(a), a typical Mask R-CNN model, trained on the 20 VOC classes from the COCO dataset, effectively identifies objects within the training taxonomy such as people and chairs. However, it struggles to detect objects beyond this taxonomy, like hats, paper, and plates.

Conversely, classical bottom-up segmentation methods~\citep{uijlings2013selective, pont2016multiscale, grundmann2010efficient} are class-agnostic and unsupervised by design, making them suitable for open-world scenarios. These methods rely solely on low-level cues such as shape, size, color and texture to generate object masks. However, they often suffer from over-segmentation, lacking a semantic understanding of objectness. In \figref{fig:intro}(b), MCG~\citep{pont2016multiscale} generates over-segmentation of objects with noisy boundaries.


\review{

How can we combine the strengths of both paradigms? We answer this question with our novel approach for open-world instance segmentation, termed \Ours{} (Bottom-Up and Top-Down Open-World Segmentation). \Ours{} seamlessly integrates the advantages of both top-down and bottom-up methods into a unified and jointly trainable framework. \Ours{} (\figref{fig:intro}c) effectively segments known categories like persons and chairs while demonstrating robust generalization to unseen categories like party hats, paper, glasses, and plates.

}
%

%
\Ours{} is grounded on two key intuitions:
\review{First, we recognize the value of weak supervision from class-agnostic segmentation generated by unsupervised bottom-up methods~\citep{uijlings2013selective, pont2016multiscale, grundmann2010efficient}. This supplementation complements potentially incomplete human annotations, ensuring holistic image segmentation without designating any region as negative. Second, we leverage seen-class supervision to bootstrap objectness, introducing an affinity-based grouping module to merge parts into complete objects and a refinement module to enhance boundary quality. Despite only being trained on seen categories, we observe that both part-level grouping and boundary refinement generalize well to novel categories.

\Ours{} is the first approach that effectively combines top-down architecture and bottom-up supervision into a unified framework for open-world instance segmentation, and we show its superiority over existing methods 
through extensive empirical experiments. Our contributions are:

}
%
\begin{tight_enumerate}
    \item We propose \Ours{} for open-world instance segmentation that effectively combines bottom-up unsupervised grouping with top-down learning in a single jointly trainable framework (\secref{sec:method}). 
    \item We propose an affinity based grouping strategy (\secref{subsec:grouping}) followed by a refinement module (\secref{subsec:refinement}) to convert noisy part-segmentations into coherent object segmentations. We show that such grouping generalizes well to unseen objects.
    \item \Ours{} achieves significant improvements over competitive baselines as well as recent open-world instance segmentation methods OLN~\citep{kim2022learning}, LDET~\citep{LDET} and GGN~\citep{wang2022open} on cross-category generalization (VOC to NonVOC)
    as well as cross-dataset (COCO to UVO,  ADE20K and OpenImagesV6) settings (\secref{sec:expmnts}).
\end{tight_enumerate}



\section{Related Works}

\begin{figure*}[!t]
\begin{center}
    \begin{minipage}[t]{0.42\textwidth}
        \centering
        \includegraphics[width=\textwidth]{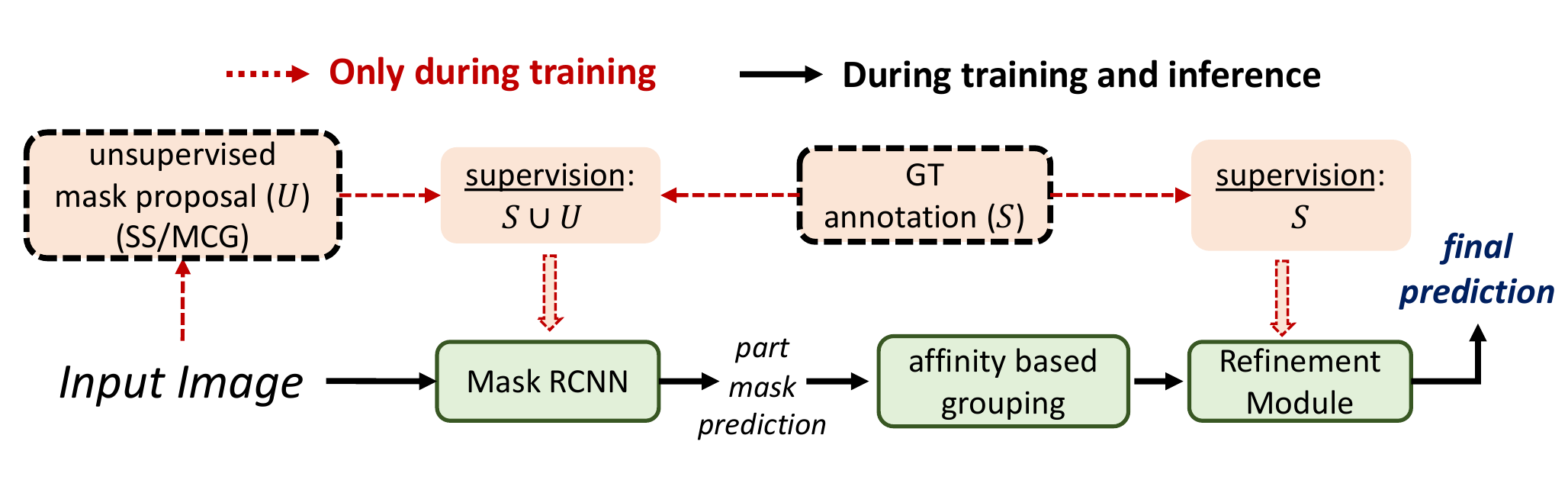}
        \captionsetup{width=\textwidth, font=footnotesize}
        \caption{\Ours{} overview}
        \label{fig:udos_overview}
    \end{minipage}
    \hfill
    \begin{minipage}[t]{0.56\textwidth}
        \centering
        \includegraphics[width=\textwidth]{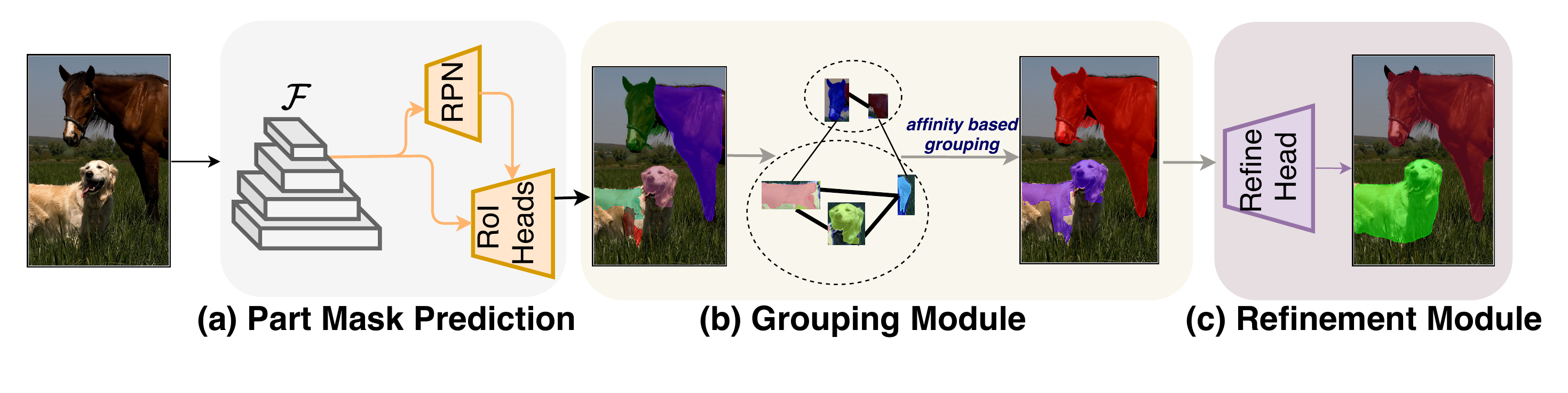}
        \captionsetup{width=\textwidth, font=footnotesize}
        \caption{Proposed \Ours{} pipeline}
        \label{fig:overview}
    \end{minipage}
    \caption*{{\bf (\figref{fig:udos_overview}) \Ours{} overview:} \review{Training and inference phases in \Ours{}}. The unsupervised proposal generation is only present during training and not used in inference.  {\bf (\figref{fig:overview}) Proposed \Ours{} pipeline:} During training, we first augment the ground truth annotations on seen classes ($S$) with masks provided by the unsupervised segmentation algorithm ($U$) and use it to supervise the part mask prediction head in (a) (\secref{subsec:part_mask_predict}). As these predictions might only correspond to part-segments on unknown classes (\textit{head} of the horse, \textit{body} of the dog), we use an affinity based grouping strategy in (b) that merges part segments corresponding to the same instance (\secref{subsec:grouping} and \figref{fig:grouping}). We then apply a refinement head in (c) to predict high-quality segmentation for complete instances.
        }
\end{center}
\end{figure*}
    
\noindent \textbf{Object detection and instance segmentation.} In the past, these tasks relied on handcrafted low-level cues with a bottom-up approach: graph-based grouping~\citep{gb,GBH}, graph-based methods~\citep{ShiM00,BoykovVZ01,felzenszwalb2008discriminatively}, deformable parts~\citep{felzenszwalb2009object}, hierarchical and combinatorial grouping~\citep{ucm,pont2016multiscale} or Selective Search~\citep{uijlings2013selective}. The rise of deep learning brought about top-down approaches, excelling in various detection and segmentation tasks, including object proposals~\citep{pedro2015learning,EdgeBox}, object detection~\citep{FasterRCNN}, semantic segmentation~\citep{fcn}, instance segmentation~\citep{he2017mask,arnab2017pixelwise,bolya2019yolact} and panoptic segmentation~\citep{KirillovHGRD19,MaX-DeepLab}. However, this paper addresses a fundamentally different challenge. Instead of assuming a closed-world scenario where training and testing share the same taxonomy, we focus on the open-world, which involves segmenting both in-taxonomy and out-of-taxonomy instances. As in \figref{fig:intro} and \secref{subsec:crosscategory}, top-down methods exhibit a bias towards seen classes and struggle to detect novel objects.

\textbf{Open world instance segmentation.} Open-world vision involves generalizing to unseen objects~\citep{bendale2015towards,pedro2015learning} and is regaining traction in computer vision~\citep{hsu2020generalized,liu2019large,wang2021unidentified,abs-2104-11221,joseph2021open,kim2022learning}. We focus on open-world instance segmentation~\citep{wang2021unidentified}, where the goal is to detect and segment objects, even if their categories weren't in the training data. This differs from works~\citep{hu2018learning,kuo2019shapemask} that rely on categories with bounding box annotations during training. Prior methods~\citep{du2021learning, pham2018bayesian, dave2019towards, jain2017fusionseg} often used additional cues like video, depth, or optical flow. In contrast, \Ours{} requires no extra annotations and relies on unsupervised proposal generation. Our work is related to ~\citep{wang2021unidentified,kim2022learning,wang2022open}. \citet{wang2021unidentified} introduces a benchmark, while our paper presents a novel approach. OLN~\citep{kim2022learning} enhances objectness but uses only seen-class annotations in training. \Ours{} combines top-down training and bottom-up grouping for novel object segmentation. GGN~\citep{wang2022open} shares a similar approach with \Ours{} by using bottom-up grouping. However, GGN uses pixel-level pairwise affinities for grouping, while \Ours{} uses part-level pairwise affinities, with fundamentally different grouping principles. Our results show that \Ours{} performs well compared to GGN, indicating it could be a complementary method. Finally, \Ours{} is also complimentary to recent innovations like Segment Anything (SAM) \citep{kirillov2023segment}, where the potential use of masks generated by SAM as initial segmentations can further improve the quality of open-world detections by \Ours{}.

\noindent \textbf{Combining bottom-up and top-down.} Recent research has revisited bottom-up methods in representation learning~\citep{henaff2021efficient, bar2021detreg, zhang2020self}. In instance segmentation, bottom-up grouping has improved local segmentation quality using affinity maps~\citep{liu2018affinity, bailoni2019generalized, xu2020deep,wang2022open} or pixel grouping~\citep{liu2017sgn, hwang2019segsort, siam2021video}. However, these approaches focus on closed-world taxonomies. Our work combines top-down training and bottom-up grouping for open-world instance segmentation, distinguishing it from prior grouping-based methods~\citep{kong2018recurrent, ahn2018learning} that use low-level pixel features. We also address open-world instance segmentation, unlike prior work on 3D part discovery or shape analysis~\citep{luo2020learning}.




\section{Proposed Method}
\label{sec:method}

\begin{figure*}[!t]
    \centering
    \includegraphics[width=0.75\textwidth]{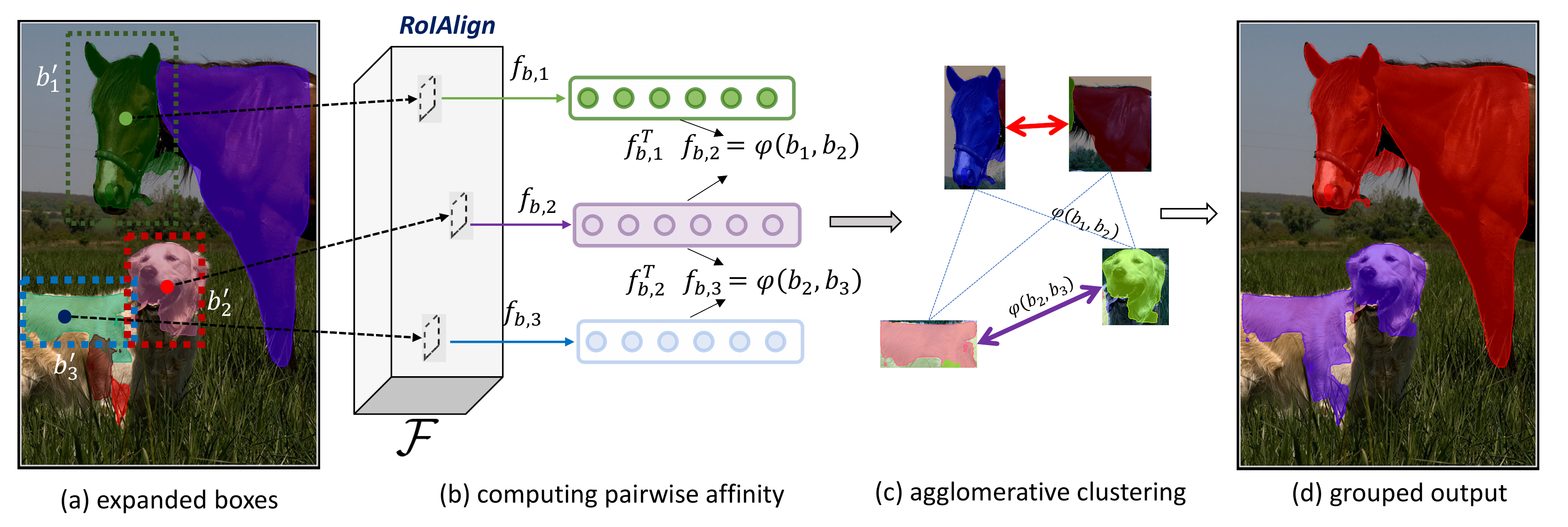}
    \caption{{\bf Grouping module}. (a) the bounding boxes $b_i$ of the predicted part-masks are expanded to incorporate local context. (b) The features $f_{b,i}$ are extracted using RoIAlign operator on the FPN features $\mathcal{F}$ with the expanded bounding boxes $b_i'$, and are used to compute pairwise affinity $\phi(b_i,b_j)$ using cosine similarity. (c) A clustering algorithm is used to group parts into whole object instances, as shown in (d). Note that the inaccuracies in the output from grouping module are later corrected by the refinement module.}
    \label{fig:grouping}
\end{figure*}

\noindent {\bf Problem definition}. Given an image $I {\in} \mathbb{R}^{H {\times} W {\times} 3}$, the goal of open world instance segmentation is to segment all object instances in $I$ regardless of their semantic categories, which includes objects that were both seen and unseen during training. Following prior works~\citep{pedro2015learning, kim2022learning, wang2022open}, we adopt class-agnostic learning strategy, in which all annotated classes are mapped to a single foreground class during training and predictions are class-agnostic.

\noindent {\bf Method overview of \Ours{}}. We visualize the training and inference flows of \Ours{} in \figref{fig:udos_overview}. \Ours{} consists of part-mask prediction (\figref{fig:overview}a), affinity-based grouping (\figref{fig:overview}b) and refinement (\figref{fig:overview}c). We use class-agnostic Mask R-CNN~\citep{he2017mask} with FPN~\citep{lin2017feature} as backbone, and we denote the FPN feature map as $\mathcal{F}$. 

\subsection{Part-Mask Prediction}
\label{subsec:part_mask_predict}

\noindent {\bf Generating candidate object regions}. We start by creating weak supervision using unsupervised segmentation algorithms (e.g., selective search~\citep{uijlings2013selective} or MCG~\citep{pont2016multiscale}) for each image in the training set. These segmentation masks are class-agnostic and cover the entire image, regardless of in-taxonomy or out-of-taxonomy objects. We intentionally favor over-segmentation during proposal generation by tuning the algorithms' hyperparameters (e.g., scale and $\sigma$ in selective search). It's important to note that this process is a \emph{one-time} effort before training and is not needed during inference (\figref{fig:overview}).


\noindent {\bf Augmenting labels using part-masks}. Next, for each training image $I$, we create a triplet ($I, S, U$), where $S = \{s_i\}_{i=1}^{N_s}$ represents the set of ground truth box and mask labels for annotated categories, and $U = \{u_i\}_{i=1}^{N_u}$ represents masks generated by the unsupervised segmentation algorithm, offering more extensive but potentially noisy region proposals. We use the augmented masks set $A {=} S {\cup} U$ as supervision to train a top-down instance segmentation system, referred to as the part-mask prediction network. This network may predict only parts of objects in alignment with the provided supervision (output \figref{fig:overview}a). To avoid label duplication, we exclude part masks from $U$ that overlap with any ground truth mask in $S$ with an IoU greater than 0.9. In essence, while masks in $S$ provide information for detecting in-taxonomy classes, masks in $U$ assist in segmenting part masks for \textbf{all} objects, offering complementary training signals to the network. This strategy offers two key advantages over top-down training with only ground truth masks in $S$. First, unsupervised region proposals from $U$ account for un-annotated image regions that may contain valid out-of-taxonomy objects, preventing the network from mistakenly labeling them as background. Second, despite masks in $U$ potentially not representing complete objects, they still provide valuable training signals for detecting out-of-taxonomy objects. For example, accurately detecting parts of a dog, such as the head, body, and ears in \figref{fig:overview}, proves useful in the final segmentation of the entire dog through our part-mask grouping strategy.

\subsection{Grouping Module}
\label{subsec:grouping}

To bridge the gap between mid-level part-masks (\secref{subsec:part_mask_predict}) and complete object instances, we propose an efficient lightweight grouping strategy to merge parts into objects. We compute pairwise affinities between features of the \textit{expanded} parts, and cluster them based on affinities. 

\noindent {\bf Pairwise affinity} We denote the predictions made by the network in the first phase by $P = \{p_i\}_{i=1}^{n_p}$, where $n_p$ is the number of predictions, and $p_i$ contains mask ($m_i$) and box ($b_i$) predictions made on seen as well as unseen categories.
For each bounding box $b_i {\in} p_i$, we first expand the width and height of the box by a factor $\delta (0 {<} \delta {<} 1)$ to compute a new, larger bounding box $b_i'$ (\figref{fig:grouping}a). 
\begin{equation}
    b_i : (x_i,y_i,h_i,w_i) \xrightarrow{\text{expand}} b_i' :  (x_i,y_i,(1+{\delta}){*}h_i,(1+{\delta}){*}w_i) 
    \label{eq:expansion}
\end{equation}
where $(x_i,y_i)$ is the center and $(h_i,w_i)$ are the original height and width of box $b_i$. This inflation allows us to ingest useful context information around the part and the underlying object, better informing the affinities between the part-masks.
Next, we compute the ROIAlign features for all the boxes $\{b_i'\}$ from the FPN feature map $\mathcal{F}$ resulting in a $d$-dim feature for each part-prediction denoted using $\{f_{b,i}\}_{i=1}^{n_p} {\in} \mathbb{R}^d$. 
The pairwise affinity between two part-predictions $(p_i,p_j) \in P$ is then computed using the cosine similarity between the corresponding feature maps (\figref{fig:grouping}b).
\begin{equation}
    \phi(p_i, p_j) = \frac{ {f_{b,i}}^T \cdot f_{b,j} }{ \|f_{b,i}\| \|f_{b,j}\| }; f_{b,i} = \text{RoIAlign}(\mathcal{F}, b_i') 
    \label{eq:affinity}
\end{equation}
%
We visualize the parts retrieved using pairwise affinity for few examples in \figref{fig:affinitymaps}. 
While \citet{wang2022open} has shown strong generalization of pixel pairwise affinities, our novelty lies in showing that the part-mask pairwise affinities generalize better across object categories.

\vspace{2pt}
\noindent {\bf Affinity based grouping} We use a clustering algorithm to merge parts based on the soft affinity scores given in \eqref{eq:affinity}. 
Our clustering objective can be formulated as follows:
\begin{eqnarray}
    \max_{G} \sum_{k=1}^{|G|} \sum_{p_i, p_j \in g_k} \phi(p_i, p_j), \quad
    \text{s.t.} \sum_{k=1}^{|G|} |g_k| = n_p
\end{eqnarray}

\noindent where $G$ is a possible partition of the $n_p$ predictions, $|G|$ denotes the total number of partitions and $k^{\text{th}}$ partition in $G$ is denoted by $g_k$ ($1 {\leq} k {\leq} |G|$). 
In other words, given a set of elements along with their pairwise affinity scores, our clustering algorithm produces a partition of the elements that maximizes the average affinities within each partition. We use an off-the-shelf agglomerative clustering algorithm from~\citet{bansal2004correlation} provided by \texttt{scikit-learn}~\citep{scikit-learn}. It is parameter-free, lightweight, and fast, incurring minimum time and memory overhead while clustering hundreds of part-masks in each iteration. As shown in \secref{sec:ablations} our final framework adds negligible inference time overhead to the backbone. We merge all the part masks (and boxes) within each partition group $g_k$ to form more complete masks, representing whole objects (\figref{fig:overview}b). Since the original predictions in $P$ might also represent whole objects on seen classes, we combine the originally detected masks as well as the grouped masks into our output at this stage. 

\subsection{Refinement Module}
\label{subsec:refinement}

\looseness=-1

To address potential blurriness in the grouped masks due to noisy initial segmentation, we incorporate a refinement module. Designed similar to the RoIHeads in Mask R-CNN, this module takes the predictions generated after the grouping stage as inputs (\figref{fig:overview}c). We train the refinement head exclusively on annotated ground truth instances from $S$ to introduce the concept of object boundaries into the predictions (only available in the annotated masks). We found that this boundary refinement also generalizes well to unseen categories. We jointly train the backbone and refinement heads in a single stage, using losses from the part-mask prediction and refinement modules.

\noindent {\bf Objectness ranking}
Following \citep{kim2022learning}, we add box and mask IoU branches to our RoIHead in part-mask predictions as well as refinement heads to compute the localization quality. IoU metrics are shown to improve objectness prediction~\citep{huang2019mask} and avoid over-fitting to seen instances~\citep{kim2022learning} when trained with non-exhaustive annotations. 
We use box and mask IoU heads with two fc-layers of 256-dim each followed by a linear layer to predict the IoU score, trained using an L1 loss for IoU regression. 

\noindent{\bf Inference} (\figref{fig:udos_overview}).
We first predict part masks, followed by the affinity based grouping to hierarchically merge them into complete objects. We then pass these detections through the refinement layer to obtain the final predictions. We rank the predictions using the geometric mean of their predicted classification score $c$, box IoU $b$ and mask IoU $m$ 
from the refinement head as $s = \sqrt[3]{c * b * m}$. 


\section{Experiments}
\label{sec:expmnts}

\looseness=-1
\noindent {\bf Datasets and evaluations.}
We demonstrate the effectiveness of \Ours{} for open-world instance segmentation under \textit{cross-category} generalization within the same dataset, as well as \textit{cross-dataset} generalization across datasets with different taxonomies (\tabref{tab:number_of_inst}).
We use the MS-COCO~\citep{lin2014microsoft} for cross-category generalization and train the model using 20 categories from VOC and test on the 60 remaining unseen nonVOC classes following prior work~\citep{kim2022learning, LDET, wang2022open}. For cross-dataset generalization, we train on complete COCO dataset and directly test on validation splits of UVO~\citep{wang2021unidentified}, ADE20k~\citep{zhou2019semantic} and OpenImagesV6~\citep{benenson2019large} datasets without any fine-tuning. We also test large-taxonomy scenario by training on a subset of 1123 categories from LVIS~\citep{gupta2019lvis} and test on COCO. Both UVO and ADE20k datasets provide exhaustive annotations in every frame, which is ideal to evaluate open world models, while OpenImagesV6 with 350 categories allows to test our open world segmentation approach on large scale datasets.


\begin{table}[!t]
  \centering
  \begin{minipage}[t]{0.86\hsize}\centering
  \resizebox{\textwidth}{!}{
  \begin{tabular}{@{} l *{5}{c} @{}} 
    \toprule  \\[-1em]
    \multicolumn{4}{c}{\textbf{Cross-category setting}} \\
    \midrule
    Train On & Test On & \# Seen classes & \# Unseen classes  \\
    \midrule
    VOC & Non-VOC & 20 & 60 \\
    LVIS & COCO & 1123 & 80 \\
    \bottomrule 
  \end{tabular}
  }
    \end{minipage}
    
    \begin{minipage}[t]{0.86\hsize}\centering
  \resizebox{\textwidth}{!}{
  \begin{tabular}{@{} l *{5}{c} @{}} 
    \toprule  \\[-1em]
    \multicolumn{4}{c}{\textbf{Cross-dataset setting}} \\
    \midrule
    Train On & Test On & \# Seen classes & \# Unseen classes  \\
    \midrule
    \multirow{3}{*}{COCO} & UVO & \multirow{3}{*}{80} & open \\
    & ADE20k & & 70 \\
    & OpenImagesV6 &  & 270 \\
    \bottomrule 
  \end{tabular}
  }
    \end{minipage}
  \caption{{\bf Evaluation settings}. Seen and unseen categories used in our evaluation.}
    \label{tab:number_of_inst}
\end{table}



\begin{figure*}[!tbp]
\begin{center}
        \centering
        \includegraphics[width=\textwidth]{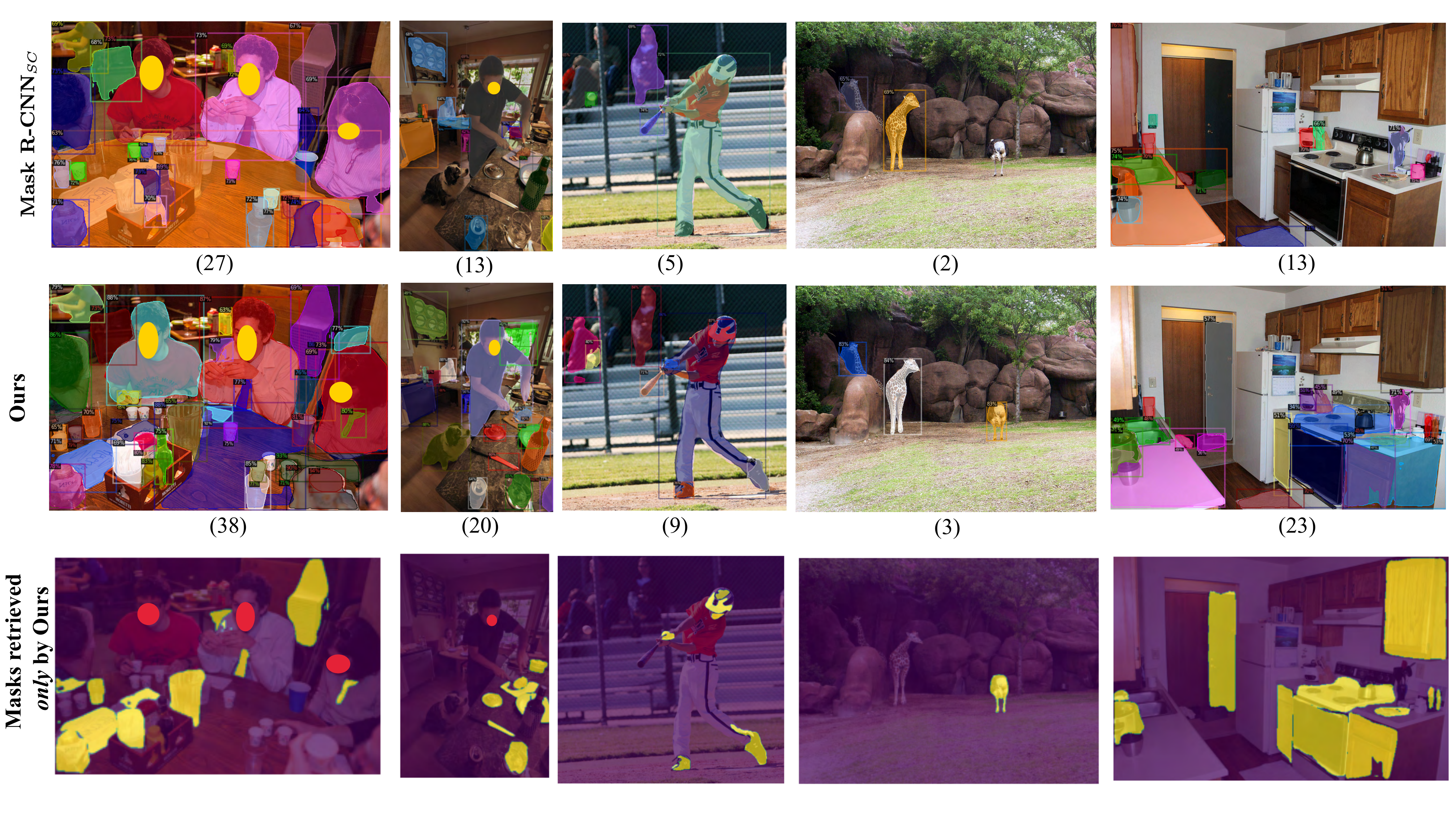}
        \captionsetup{width=\textwidth, font=footnotesize}
        \caption{{\bf Visualization of segmentations for model trained only on VOC classes} from COCO dataset. The top row shows result using using Mask-RCNN$_{SC}$, second row shows output using \Ours{} and the third row shows some predictions made only by \Ours{} and missed by Mask-RCNN$_{SC}$ . We also show the number of detections made by the network below each image. Starting from left most image, many classes like \{jug, tissue papers, tie, eyeglasses\}, \{knife, cutting board, vegetables, glass\}, \{shoes, helmet, gloves\}, \{ostrich\} and \{dishwasher, faucet\} among others which are not part of VOC-classes are missed by standard Mask-RCNN training, but detected using \Ours{}. More visualizations are provided in the \supp{}.
        }
        \label{fig:visualization}
\end{center}
\end{figure*}

\noindent {\bf Implementation details.}
We use Mask R-CNN model~\citep{he2017mask} with a ResNet-50-FPN~\citep{lin2017feature} as our backbone. We train \Ours{} using SGD for $10$ epochs with an initial learning rate of $0.02$ on $8$ GPUs. We use selective search~\citep{uijlings2013selective} to generate unsupervised masks for images in COCO dataset.
Note that the mask proposals are required \emph{only during training}, and \emph{not} during inference (\figref{fig:udos_overview}). We follow prior works in open-world instance segmentation~\citep{kim2022learning, LDET, wang2022open} and use average recall (AR) (between IoU thresholds of 0.5 to 1.0) as the evaluation metric. 
Since open world models generally detect many more objects in a scene than closed world models (see \figref{fig:visualization}) and many datasets do not have exhaustive annotation, we use AR$^{100}$ and AR$^{300}$ as the evaluation metrics on both box (AR$_B$) and mask (AR$_M$) to avoid penalizing predictions of valid, yet unannotated, objects.

\subsection{Baselines}
\begin{enumerate*}[label=(\roman*)]
    \item{\bf Image-computable masks}: We use masks generated by {\bf MCG}~\citep{pont2016multiscale} and Selective Search~\citep{uijlings2013selective} ({\bf SS}), which are class-agnostic, learning-free proposal generation methods relying on low-level cues. 
    \item{\bf Mask-RCNN}~\citep{he2017mask} denotes Mask R-CNN training in class-agnostic fashion only on the seen classes,
    \item{\bf Mask R-CNN$_{SS}$} indicates Mask R-CNN trained using selective search proposals as the supervision instead of the ground truth annotations, and 
    %
    \item{\bf Mask R-CNN$_{SC}$} denotes Mask R-CNN trained with BoxIoU and MaskIoU scoring to rank the proposals instead of the classification score. 
    %
    %
\end{enumerate*}

\begin{table}[!t]
\begin{minipage}[t]{0.94\hsize}\centering
\centering
      \resizebox{\textwidth}{!}{
      \begin{tabular}[t]{@{} l *{12}{c} @{}} 
            \toprule 
            \textit{VOC${\rightarrow}$NonVOC}& AR$_B^{100}$ & AR$_B^{300}$ && AR$_M^{100}$ & AR$_M^{300}$ \\
            \cmidrule{2-3} \cmidrule{5-6}
            \multicolumn{6}{l}{\textit{\textbf{Bottom-up}(No Training)}} \\
            \cline{1-6}
            SS & 14.3 & 24.7 && 6.7 & 12.9 \\
            MCG & 23.6 & 30.8 && 19.4 & 25.2 \\
            \multicolumn{6}{l}{\textit{\textbf{Top-down}(Class-agnostic Training)}} \\
            \cline{1-6}
            MaskRCNN & 25.1 & 30.8 && 20.5 & 25.1 \\
            Mask R-CNN$_{SS}$ & 24.1 & 24.9 && 20.9 & 21.7 \\
            {Mask R-CNN$_{SC}$} & 25.6 & 33.1 && 24.9 & 28 \\
            \multicolumn{6}{l}{\textit{\textbf{Open-World Methods}}} \\
            \cline{1-6}
            OLN & \noop{32.5} & \noop{37.4} && \noop{26.9} & \noop{30.4} \\
            LDET & 30.9 & 38.0 && 26.7 & 32.5 \\
            GGN & 31.6 & 39.5 && 28.7 & 35.5 \\
            \cline{1-6} \\ [-1em]
            \Ours{} & \textbf{33.5} & \textbf{41.6} && \textbf{31.6} & \textbf{35.6} \\
            \bottomrule
        \end{tabular}
        }
    \captionsetup{width=\textwidth, font=footnotesize}
    \caption{{\bf Cross-category generalization evaluation on COCO.} Train on 20 VOC categories and test on 60 NonVOC categories. \Ours{} outperforms many competitive baselines as well as the {current SOTA GGN} on the VOC${\rightarrow}$NonVOC setting.
    }
    \label{tab:voc2nvoc}
    \vspace{-16pt}
    \hfill
    \end{minipage}
\end{table}

We also compare with state of the art open-world instance segmentation algorithms OLN~\citep{kim2022learning}, LDET~\citep{LDET} and GGN~\citep{wang2022open}. For fair comparison with \Ours{}, we use the result from GGN~\citep{wang2022open} \textit{without} the OLN backbone.

\subsection{\Ours{} outperforms baselines on cross-category generalization} 
\label{subsec:crosscategory}


Existing methods relying on bottom-up grouping with techniques like SS or MCG, or top-down architectures like Mask R-CNN trained on annotations for seen classes, struggle to effectively detect and segment instances from unseen classes. In contrast, \Ours{} overcomes these limitations (\tabref{tab:voc2nvoc}) by harnessing both ground truth annotations for known classes and unsupervised bottom-up masks for unknown classes, resulting in a significant improvement over all baseline methods, underscoring the effectiveness of our approach. From \tabref{tab:voc2nvoc}, \Ours{} achieves Box Average Recall (AR) of 33.5\% and Mask AR of 31.6\% setting a new state-of-the-art in cross-category open-world instance segmentation, outperforming the current state-of-the-art methods like GGN in both box and mask AR. 


Expanding training datasets to encompass larger taxonomies is a potential strategy for addressing the challenge of novel categories at test-time. However, our experiments, detailed in \tabref{tab:lvis}, reveal that this approach still falls short of achieving robust generalization to unseen categories. We leveraged the LVIS dataset \citep{gupta2019lvis}, which includes annotations for 1203 categories. During training, we excluded annotations with an IoU overlap greater than 0.5 with COCO masks, resulting in 79.5k instance masks from LVIS. When evaluated on COCO validation images, \Ours{} achieved 33.2\% AR$_B^{100}$ and 26.3\% AR$_M^{100}$, markedly outperforming the baseline methods. This underscores the effectiveness of \Ours{} in even handling datasets with large category vocabularies.


\subsection{\Ours{} sets new SOTA on cross-dataset generalization}
\label{subsec:crossdataset}

To provide a more realistic assessment of our model's open-world capabilities, we evaluated its performance on real-world target datasets like UVO, ADE20k, and OpenImages. These datasets contain a wide range of objects, including those not covered by COCO categories. It's worth noting that we refrained from fine-tuning our model on the target datasets or using any unlabeled target data during training. Results are in \tabref{tab:crossdataset}, and summarized below.


\begin{table}[!t]
  \centering
  \resizebox{0.45\textwidth}{!}{
  \begin{tabular}{@{} l *{7}{c} @{}} 
    \toprule  \\[-1em]
    \textit{LVIS${\rightarrow}$COCO} & AR$_B^{100}$ & AR$_B^{300}$ && AR$_M^{100}$ & AR$_M^{300}$  \\
    \midrule
    MaskRCNN & 23.8 & 29.4 && 18.5 & 22.0  \\
    Mask R-CNN$_{SC}$ & 21.3 & 27.9 && 17.9 & 24.2 \\
    OLN~\cite{kim2022learning} & {28.5} & {38.1} && {23.4} & {27.9}  \\
    \Ours{} & \textbf{33.2} & \textbf{42.2} && \textbf{26.3} & \textbf{32.2} \\
    \bottomrule
  \end{tabular}
  }
  \captionsetup{width=0.45\textwidth, font=footnotesize}
  \caption{{\bf Cross-category generalization evaluation with large taxonomy}. All models are trained on 1123 categories from LVIS (excluding COCO categories), and evaluated on COCO 80 categories. \Ours{} outperforms OLN~\cite{kim2022learning} by 4.7\% and 2.9\% on box and mask $AR^{100}$.}
    \label{tab:lvis}
    \vspace{-16pt}
\end{table}


\looseness=-1
\noindent {\bf COCO to UVO}.
Since \Ours{} is designed to handle novel classes, it achieves much better performance than other baselines on the challenging  UVO dataset that contains exhaustive annotations for every objects. \Ours{} clearly outperforms baseline approaches like Mask R-CNN (+5\% AR$_B^{100}$). We also outperform OLN, LDET and GGN, setting the new state-of-the-art on the UVO benchmark. 

\begin{table*}[!t]
  \centering
  \resizebox{\textwidth}{!}{
  \begin{tabular}{@{} l *{18}{c} @{}} 
    \toprule  \\[-1em]
    & \multicolumn{4}{c}{COCO${\rightarrow}$UVO} && \multicolumn{4}{c}{COCO${\rightarrow}$ADE20K} && \multicolumn{4}{c}{COCO${\rightarrow}$OpenImages} \\
    \cline{2-5} \cline{7-10} \cline{12-15} \\[-1em]
    & AR$_B^{100}$ & AR$_B^{300}$ & AR$_M^{100}$ & AR$_M^{300}$ && AR$_B^{100}$ & AR$_B^{300}$ & AR$_M^{100}$ & AR$_M^{300}$ && AR$_B^{100}$ & AR$_B^{300}$ & AR$_M^{100}$ & AR$_M^{300}$  \\
    \cline{2-5} \cline{7-10} \cline{12-15}
    MaskRCNN & 47.7 & 50.7 & 41.1 & 43.6 && 18.6 & 24.2 & 15.5 & 20.0 && 57.1 & 59.1 & 55.6 & 57.7  \\
    Mask R-CNN$_{SS}$ & 26.8 & 31.5 & 25.1 & 31.1 && 18.2 & 25.0 & 17 & 21.6 && 34.0 & 42.7 & 33.1 & 38.8 \\
    Mask R-CNN$_{SC}$ & 42.0 & 50.8 & 40.7 & 44.1 && 19.1 & 25.6 & 18.0 & 22.0 && 54.1 & 59.1 & 54.2 & 57.4 \\
    OLN & 50.3 & 57.1 & 41.4 & 44.7 && 24.7 & 32.1 & 20.4 & 27.2 && 60.1 & 64.1 & 60.0 & 63.5  \\
    LDET & 52.8 & 58.7 & 43.1 & 47.2 && 22.9 & 29.8 & 19.0 & 24.1 && 59.6 & 63.0 & 58.4 & 61.4 \\
    GGN & {52.8} & 58.7 & {43.4} & 47.5 && 25.3 & 32.7 & 21.0 & 26.8 && 64.5 & 67.9 & 61.4 & 64.3 \\
    \Ours{} & {\textbf{53.6}} & \textbf{61.0} & {\textbf{43.8}} & \textbf{49.2} && \textbf{27.2} & \textbf{36.0} & \textbf{23.0} & \textbf{30.2} && \textbf{71.6} &\textbf{74.6} & \textbf{66.2} & \textbf{68.7} \\
    \bottomrule
  \end{tabular}
  }
  \caption{{\bf Cross-dataset generalization evaluation for open world instance segmentation}. All models are trained on 80 COCO categories and evaluated on UVO (left), ADE20K (middle), OpenImages (right) as is without any fine-tuning. 
  }
  \label{tab:crossdataset}
  \vspace{-12pt}
\end{table*}

\noindent {\bf COCO to ADE20K} ADE20k~\citep{zhou2019semantic} is a scene parsing benchmark consisting of annotations for both \textit{stuff} (road, sky, floor etc.) and discrete \textit{thing} classes. We regard each annotation mask as a separate semantic entity and compute the average recall (AR) on both in-taxonomy and out-of-taxonomy objects to evaluate the ability of trained models to detect thing classes and group stuff classes in images. From \tabref{tab:crossdataset}, we observe that \Ours{} achieves box AR100 of 27.2\% and mask AR100 of 23.0\%, higher than all the baselines and other competing methods. 


\noindent {\bf COCO to OpenImagesV6} 
Again, \Ours{} consistently outperform all baselines as well as open-world methods like OLN and GGN by significant margins on the OpenImagesV6 dataset~\citep{benenson2019large}. We achieve AR$_B^{100}$ of 71.6\%, which is better than the strongest baseline Mask R-CNN by 14.5\% and current state-of-the-art GGN by 7.1\%. Likewise, AR$_M^{100}$ of 66.2\% obtained by \Ours{} is 4.8\% higher than GGN, setting new state of the art. 

\subsection{Ablations}
\label{sec:ablations}
We use the VOC to NonVOC cross-category generalization on COCO dataset for the ablations.

\noindent {\bf Refinement and grouping modules} We show in \tabref{tab:grpAndRefine} that without the proposed grouping and refinement modules, maskAR drops to 11.8\% from 31.6\%, as the masks are noisy and only correspond to parts of instances. Using a refinement module after grouping leads to more refined masks further improving the performance.



\noindent {\bf Choice of proposal ranking} 
We show the importance of using BoxIoU and MaskIoU scoring functions in \tabref{tab:iouscores}, where significant drops in AR100 are observed without the use of both the scoring functions, validating the observations in prior works~\citep{kim2022learning} that scoring module prevents over-fitting and improves open world learning.

\noindent {\bf Influence of $\delta$} 
Intuitively, a small value of delta (part-mask expansion factor, \eqref{eq:expansion}) would not capture sufficient context around the region for extracting similarity while a very high value of $\delta$ would induce noisy features from different neighboring objects. In \tabref{tab:delta}, we show that a value of $0.1$ achieves an optimum trade-off, so we use the same value of $\delta=0.1$ in all our experiments.


\noindent {\bf Choice of proposal generation} From \tabref{tab:spp}, we show that a naive segmentation of image using uniform grids by extracting $64{\times}64$ patches from the image expectedly performs worse, as these part masks do not semantically correspond to object parts. We also use super-pixels generated from SSN~\citep{jampani2018superpixel}, but found that bottom-up supervision generated from image-based segmentation algorithms like SS or MCG lead to much better accuracies. 




\begin{table*}[!t]
\centering
    \begin{subtable}[b]{0.24\hsize}\centering
        \resizebox{\textwidth}{!}{
        \begin{tabular}{@{} *{5}{c} @{} }
            \toprule
            Group & Refine & AR$_B^{100}$ & AR$_M^{100}$ \\
            \midrule 
            \cross & \cross & 25.4 & 11.8 \\
            \tick & \cross & 32.6 & 30.7 \\
            \tick & \tick & 33.5 &  31.6 \\
            \bottomrule
        \end{tabular}
        }
        \subcaption{}
        \label{tab:grpAndRefine}
    \end{subtable}
    \hfill
    \begin{subtable}[b]{0.24\hsize}\centering
        \resizebox{\textwidth}{!}{
        \begin{tabular}{@{} *{4}{c} @{} }
            \toprule
            BoxIoU & MaskIoU & AR$_B^{100}$ & AR$_M^{100}$ \\
            \midrule 
            \cross         & \cross         &       29.0 &       24.3               \\
            \tick           &\cross         &       32.7 &       28.9                \\
            \cross          &\tick          &       32.9 &       29.2                \\
            \tick           &\tick          &   {\bf 33.5} &     {\bf 31.6}     \\
            \bottomrule
        \end{tabular}
        }
        \subcaption{}
        \label{tab:iouscores}
    \end{subtable}
    \hfill
    \begin{minipage}[b]{0.24\hsize}\centering
        \resizebox{\textwidth}{!}{
        \begin{tabular}{@{} l *{4}{c} @{} }
            \toprule
            Segmentation & AR$_B^{100}$ & AR$_M^{100}$ \\
            \midrule 
            Uniform Grid                                        &      9.9     &     9.2       \\
            SSN                    &      19.4    &   18.7        \\
            Sel. Search            &      {\bf 33.5}    &   {\bf 31.6}        \\
            MCG                       &      32.4    &    {29.4}       \\
            \bottomrule
        \end{tabular}
        }
        \subcaption{}
        \label{tab:spp}
    \end{minipage}
    \hfill
    \begin{minipage}[b]{0.18\hsize}\centering
        \resizebox{\textwidth}{!}{
        \includegraphics[width=\textwidth]{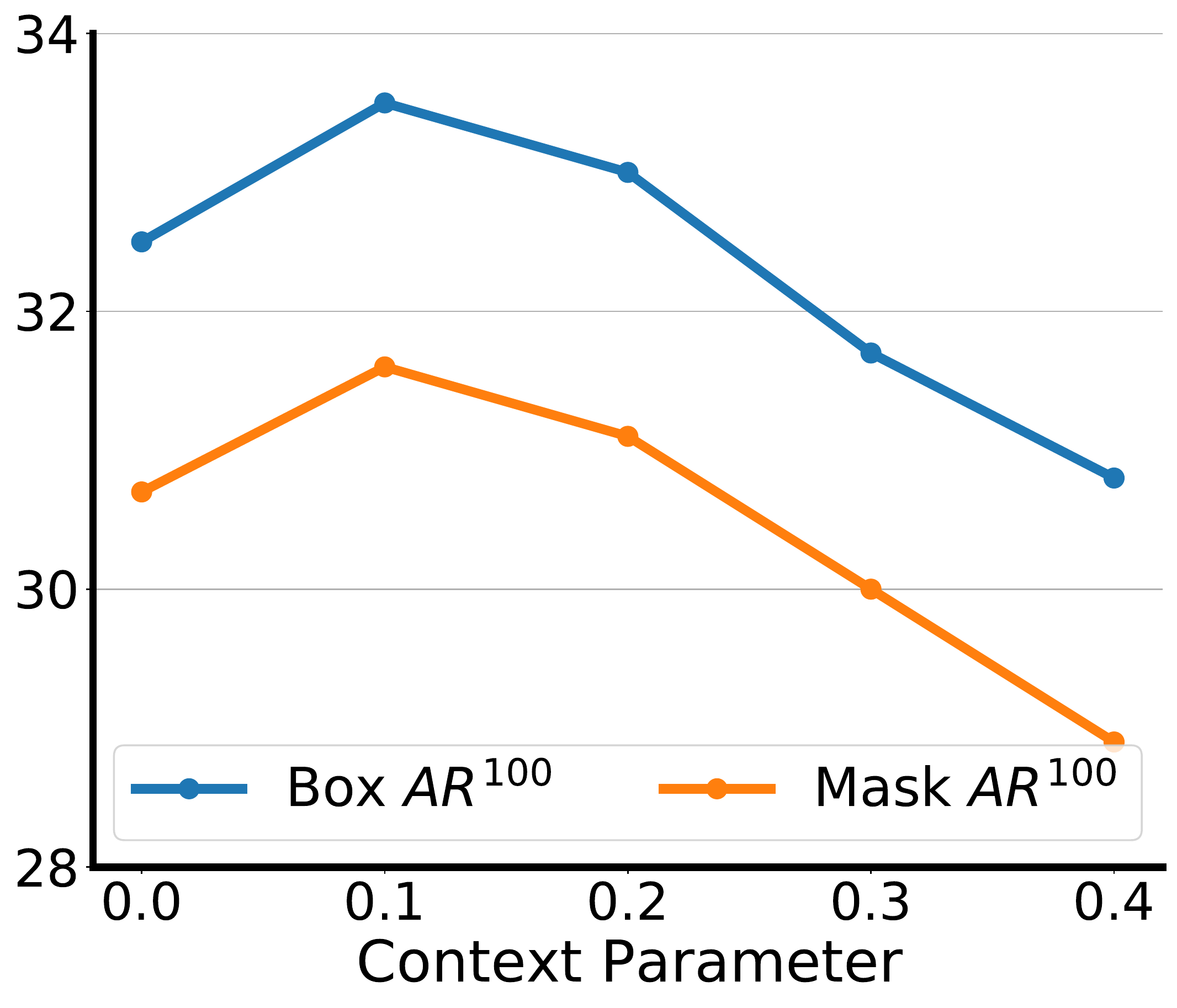}
        }
        \subcaption{}
        \label{tab:delta}
    \end{minipage}
    \vspace{-6pt}
    \captionsetup{width=\textwidth, font=footnotesize}
    \caption{{\bf Ablation results}.  Effect of (\subref{tab:grpAndRefine}) grouping and refinement modules, (\subref{tab:iouscores}), boxIoU and maskIou losses during training, (\subref{tab:spp}) segmentation algorithm and (\subref{tab:delta}) context dilation parameter $\delta$ on the VOC${\rightarrow}$NonVOC performance.  }  
    \label{tab:ablations}
\vspace{-12pt}
\end{table*}


\begin{figure*}[!tbp]
\begin{center}
        \centering
        \includegraphics[width=0.95\textwidth]{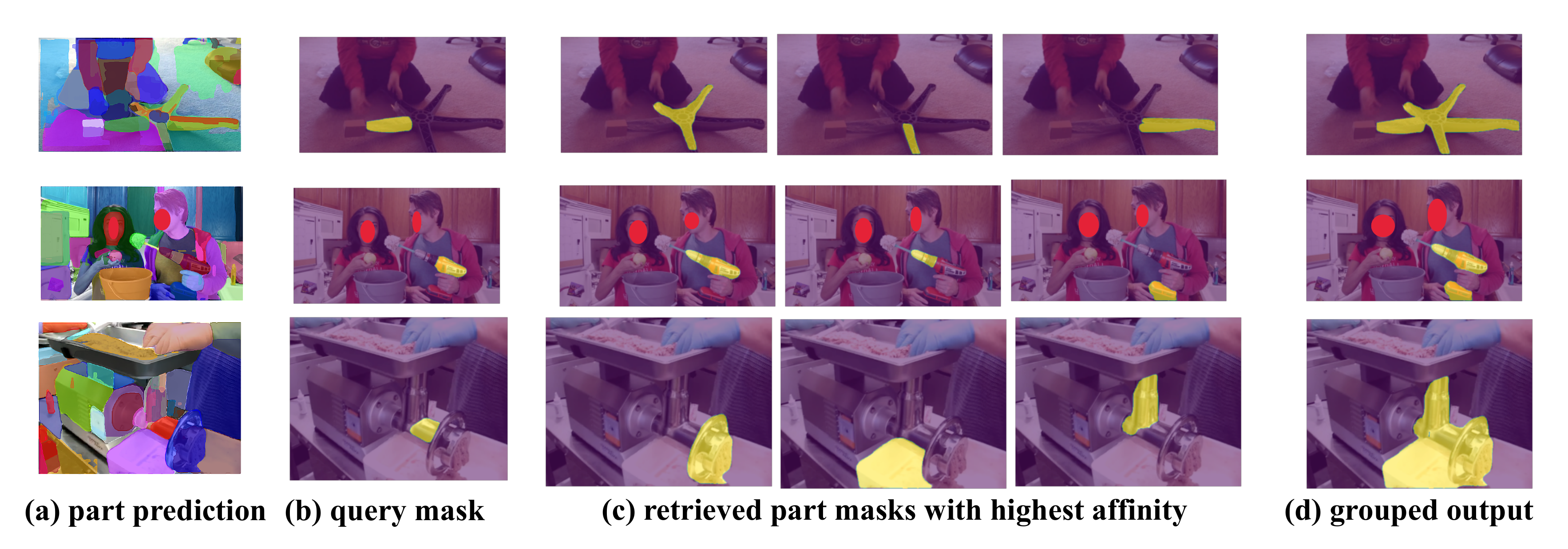}
        \captionsetup{width=\textwidth, font=footnotesize}
        \caption{{\bf Visualization of pairwise affinity maps and grouped predictions.} Given a part mask as a query, we show the 3 nearest part masks of the query using our pairwise affinity. The images are taken from UVO dataset, and the affinity is computed using \Ours{} model trained on COCO. Our affinity-based grouping module correctly groups parts into whole instances even with unseen objects. The last row visualizes a failure case where the model retrieves a part mask from a neighboring instance.
        }
        \label{fig:affinitymaps}
\end{center}
\end{figure*}

\noindent {\bf Visualizations of affinity maps} In \figref{fig:affinitymaps}, we present 3-nearest part masks retrieved for a given query mask using their affinity (\eqref{eq:affinity}) and the grouped outputs. We observe that different part masks of the same entity are often retrieved with high affinity, using our grouping module. 

\noindent {\bf Inference time comparison} 
Our grouping module is lightweight and adds negligible run-time overhead. Specifically, at 100 output proposals, MaskRCNN~\citep{he2017mask} and GGN~\citep{wang2022open} take 0.09s/im, MaskRCNN$_{SC}$ and OLN~\citep{kim2022learning} take 0.12s/im while \Ours{} takes 0.13s/im (+0.01s/im) with stronger performance. Generating part-masks using selective search for the complete COCO~\citep{lin2014microsoft} dataset takes around 1.5 days on a 16-core CPU, but we reiterate that the part-masks only need to be generated once before training and are not needed during testing/deployment (\figref{fig:udos_overview}). We will publicly release the part-masks on COCO dataset along with the code and trained models.



\section{Discussion}
\label{sec:conclusion}


In this paper, we conduct an investigation to understand what types of top-down learning generalize well in the context of open-world segmentation. Our observation revealed that learning from known classes to group part-level segmentations and learning to refine coarse boundaries are effective for generalizing to new categories. This allowed us to leverage classical bottom-up segmentation algorithms that provide class-agnostic yet coarse and over-segmented part masks within a top-down learning framework. We introduced \Ours{}, a novel approach that integrates top-down and bottom-up strategies into a unified framework. Our grouping and refinement modules efficiently convert part mask predictions into complete instance masks for both familiar and unfamiliar objects, setting \Ours{} apart from previous closed-world and open-world segmentation methods. Extensive experiments demonstrated the significant performance improvements achieved by \Ours{} across five challenging datasets, including COCO, LVIS, ADE20k, UVO, and OpenImages. We believe that \Ours{} can serve as a versatile module for enhancing the performance of downstream tasks, such as robotics and embodied AI, in handling unfamiliar objects in open-world scenarios. 


\textbf{Limitations}. \Ours{} faces challenges in scenarios with densely clustered objects of similar appearance. A more robust, learnable grouping method, possibly trained with hard negatives, could enhance performance in such complex situations, suggesting a direction for future research. Furthermore, incorporating recent innovations like Segment Anything (SAM) \citep{kirillov2023segment} to improve initial segmentations in \Ours{} is also en exiciting future directions to follow.

{
    \small
    \bibliographystyle{ieeenat_fullname}
    \bibliography{main}
}

\appendix

\begin{table*}[!!h]
\centering
    \captionsetup{width=\textwidth, font=footnotesize}
    \caption{{\bf Effect of weight sharing between RoI head and refinement head.} Comparison of results with and without sharing parameter weights between RoI module of part mask prediction head and the refinement head. Using separate heads for the RoI head of part-mask prediction module and refinement module improves AR$_M^{100}$ by $3.3\%$.
    }
    \label{tab:weightsharing}
      \resizebox{0.9\textwidth}{!}{
      \begin{tabular}{@{} l *{18}{c} @{}} 
            \toprule 
            & AR$_B^{100}$ & AR$_B^{300}$ & AR$_B^{s}$ & AR$_B^{m}$ & AR$_B^{l}$ && AR$_M^{100}$ & AR$_M^{300}$ & AR$_M^{s}$ & AR$_M^{m}$ & AR$_M^{l}$ \\
            \midrule
            Shared weights & {33.5} & {41.3} & 13.2 & {43.6} & {60.6} && {28.3} & {33.6} & {11.4} & {37.7} & {48.4 }  \\
            Non shared weights & 33.5 & 41.6 & 16.2 & 43.9 & 53.5 && 31.6 & 35.6 & 15.2 & 42.7 & 48.3 \\
            \bottomrule
        \end{tabular}
        }
\end{table*}

\section{Weight sharing on refinement module}

In \Ours{}, our design of the refinement module follows the RoI-heads of Mask R-CNN architecture. Specifically, our box and mask prediction heads in the refinement head have the same architecture as the box and mask prediction heads of the Mask R-CNN. Therefore, one option is to share the weights between the RoIHeads in the first stage Mask R-CNN and the refinement heads during training. However, we note from \cref{tab:weightsharing} that such an approach of weight-sharing between the two stages of \Ours{} hurts the mask accuracy on the cross-category detection on COCO. This is because the goals of prediction in the two stages are different. While the RoIHeads in the first stage are trained to predict part-masks and trained on weak supervision from bottom-up segmentation algorithms, the refinement head is trained only using ground truth annotations and is used to predict the final object boxes and masks. However, this improvement also comes with additional increase in model parameters from 57.4M to 86.5M. Also, we observed that using individual weights only benefits cross-category setting, while cross-dataset benefits from shared weights between the part-mask MaskRCNN and refinement head. 

\section{Visualizing outputs at each stage of \Ours{}}

\begin{figure*}[!t]
    \centering
    
    \begin{subfigure}[b]{0.23\textwidth}
        \centering
        \includegraphics[width=1\textwidth]{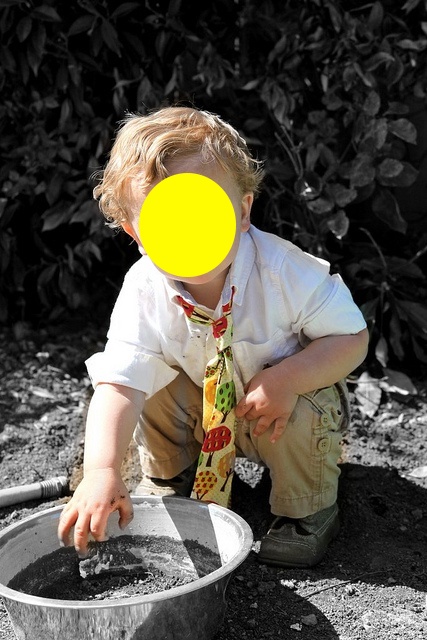}
    \end{subfigure}
    \hfill
    \begin{subfigure}[b]{0.23\textwidth}
        \centering
        \includegraphics[width=1\textwidth]{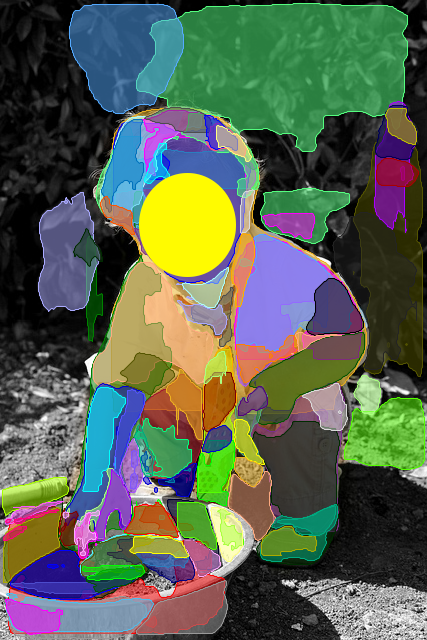}
    \end{subfigure}
    \hfill
    \begin{subfigure}[b]{0.23\textwidth}
        \centering
        \includegraphics[width=1\textwidth]{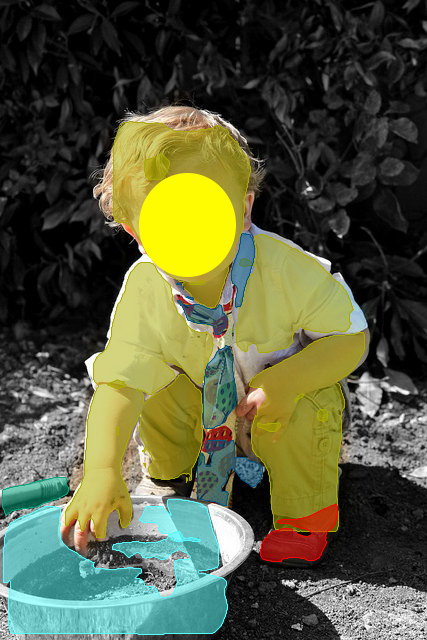}
    \end{subfigure}
    \hfill
    \begin{subfigure}[b]{0.23\textwidth}
        \centering
        \includegraphics[width=1\textwidth]{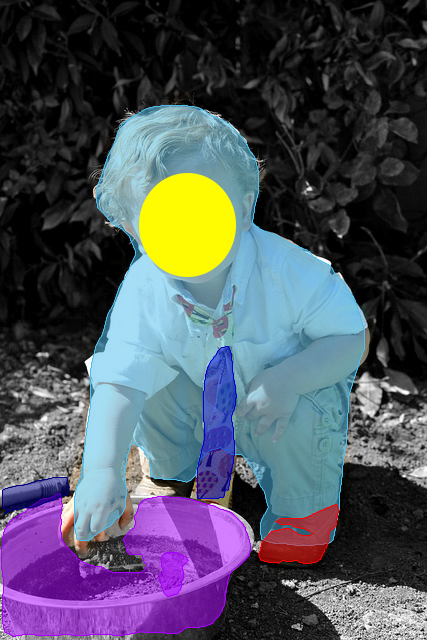}
    \end{subfigure}
    
    \vspace{4pt}
    \begin{subfigure}[b]{0.23\textwidth}
        \centering
        \includegraphics[width=1\textwidth]{figures/clustering/input_377946.jpg}
    \end{subfigure}
    \hfill
    \begin{subfigure}[b]{0.23\textwidth}
        \centering
        \includegraphics[width=1\textwidth]{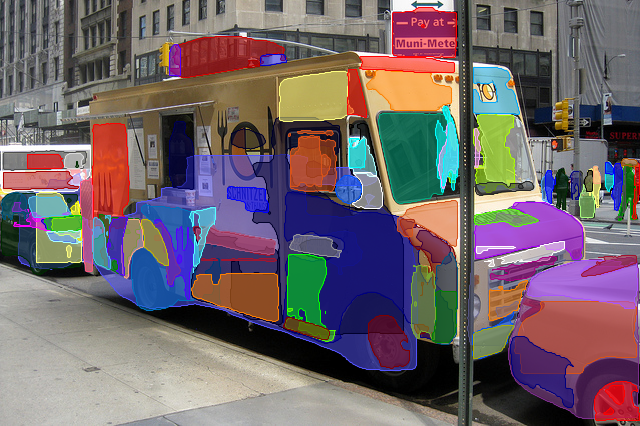}
    \end{subfigure}
    \hfill
    \begin{subfigure}[b]{0.23\textwidth}
        \centering
        \includegraphics[width=1\textwidth]{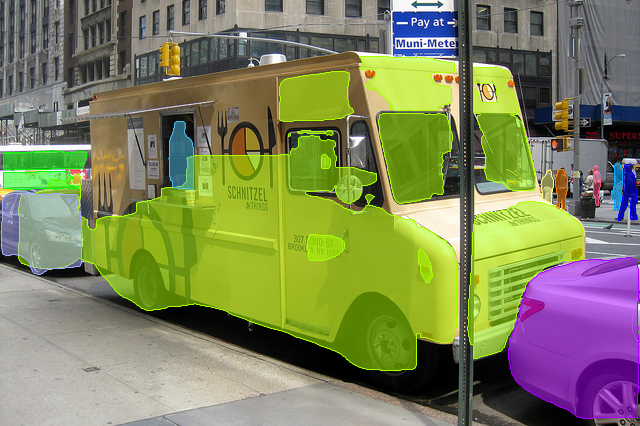}
    \end{subfigure}
    \hfill
    \begin{subfigure}[b]{0.23\textwidth}
        \centering
        \includegraphics[width=1\textwidth]{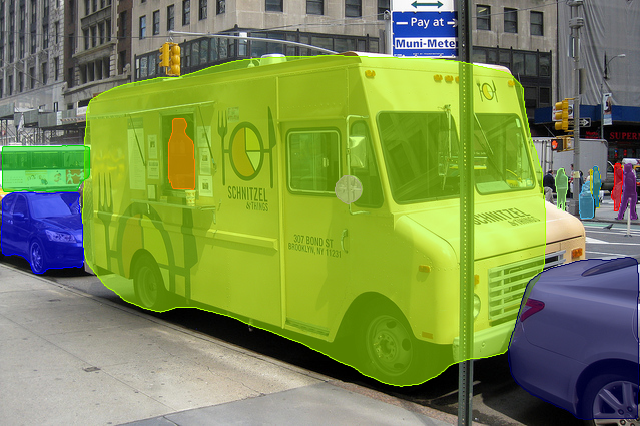}
    \end{subfigure}
    
    \vspace{4pt}
    \begin{subfigure}[b]{0.23\textwidth}
        \centering
        \includegraphics[width=1\textwidth]{figures/clustering/input_384468.jpg}
    \end{subfigure}
    \hfill
    \begin{subfigure}[b]{0.23\textwidth}
        \centering
        \includegraphics[width=1\textwidth]{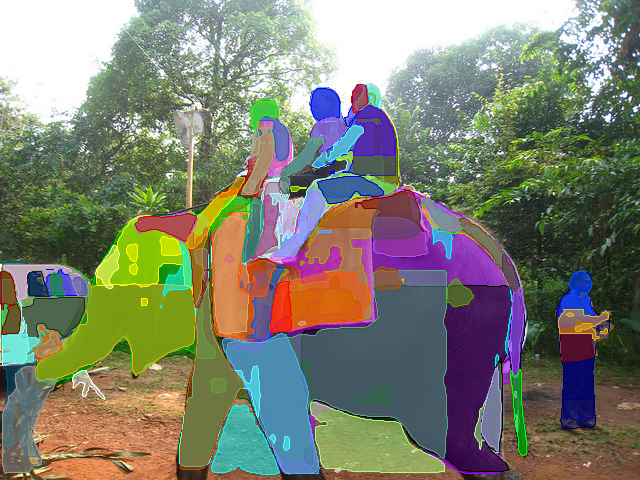}
    \end{subfigure}
    \hfill
    \begin{subfigure}[b]{0.23\textwidth}
        \centering
        \includegraphics[width=1\textwidth]{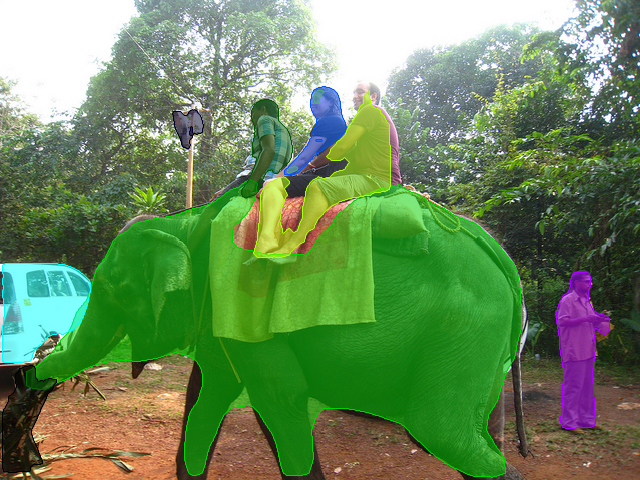}
    \end{subfigure}
    \hfill
    \begin{subfigure}[b]{0.23\textwidth}
        \centering
        \includegraphics[width=1\textwidth]{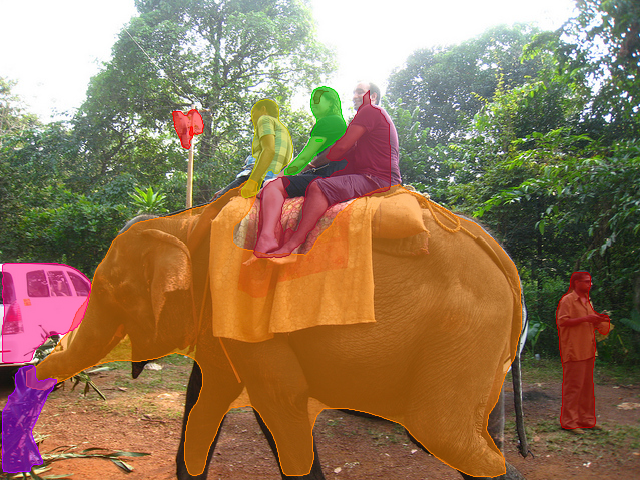}
    \end{subfigure}
    
    \vspace{4pt}
    \begin{subfigure}[b]{0.23\textwidth}
        \centering
        \includegraphics[width=1\textwidth]{figures/clustering/input_396338.jpg}
        \captionsetup{width=\textwidth, font=scriptsize}
        \subcaption{Input Image}
        \label{fig:inputimage_coco}
    \end{subfigure}
    \hfill
    \begin{subfigure}[b]{0.23\textwidth}
        \centering
        \includegraphics[width=1\textwidth]{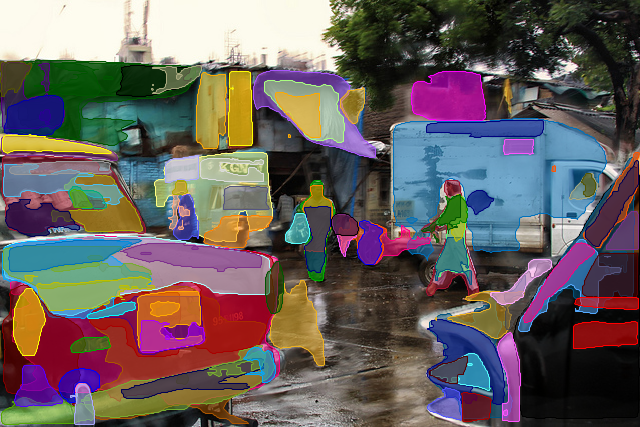}
        \captionsetup{width=\textwidth, font=scriptsize}
        \subcaption{part-mask output}
        \label{fig:partmasks_coco}
    \end{subfigure}
    \hfill
    \begin{subfigure}[b]{0.23\textwidth}
        \centering
        \includegraphics[width=1\textwidth]{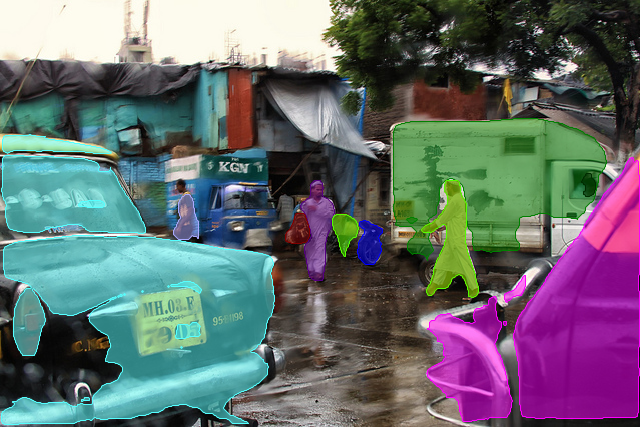}
        \captionsetup{width=\textwidth, font=scriptsize}
        \subcaption{Grouped prediction}
        \label{fig:mergedmasks_coco}
    \end{subfigure}
    \hfill
    \begin{subfigure}[b]{0.23\textwidth}
        \centering
        \includegraphics[width=1\textwidth]{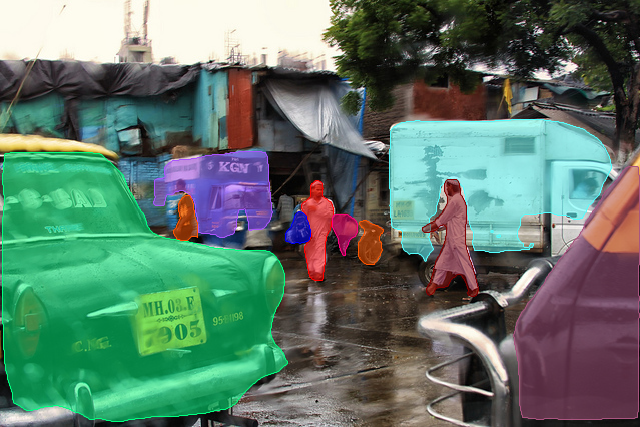}
        \captionsetup{width=\textwidth, font=scriptsize}
        \subcaption{Refined prediction}
        \label{fig:finaloutptus_coco}
    \end{subfigure}
    
    \caption{{\bf Visualizing outputs after each stage of \Ours{} for cross-category training}. All images belong to the COCO dataset, and outputs are generated using a model trained only on VOC categories. (\subref{fig:inputimage_coco}) shows the input image, followed by (\subref{fig:partmasks_coco}) Part-mask prediction, (\subref{fig:mergedmasks_coco}) grouped outputs using our affinity based grouping and (\subref{fig:finaloutptus_coco}) refined prediction. The masks in last two columns correspond to true-positives with respect to the ground truth.
    Note that classes such as \textit{pot, van, elephant}, and \textit{auto-rickshaw} do not belong to any of the training VOC categories. Also note that the merged outputs might be noisy due to the imperfection in the initial part-mask supervision used, which are corrected by our refinement layer.}
    \label{fig:cluster_voc}
\end{figure*}

\begin{figure*}[!t]
    \centering
    
    \vspace{4pt}
    \begin{subfigure}[b]{0.23\textwidth}
        \centering
        \includegraphics[width=1\textwidth]{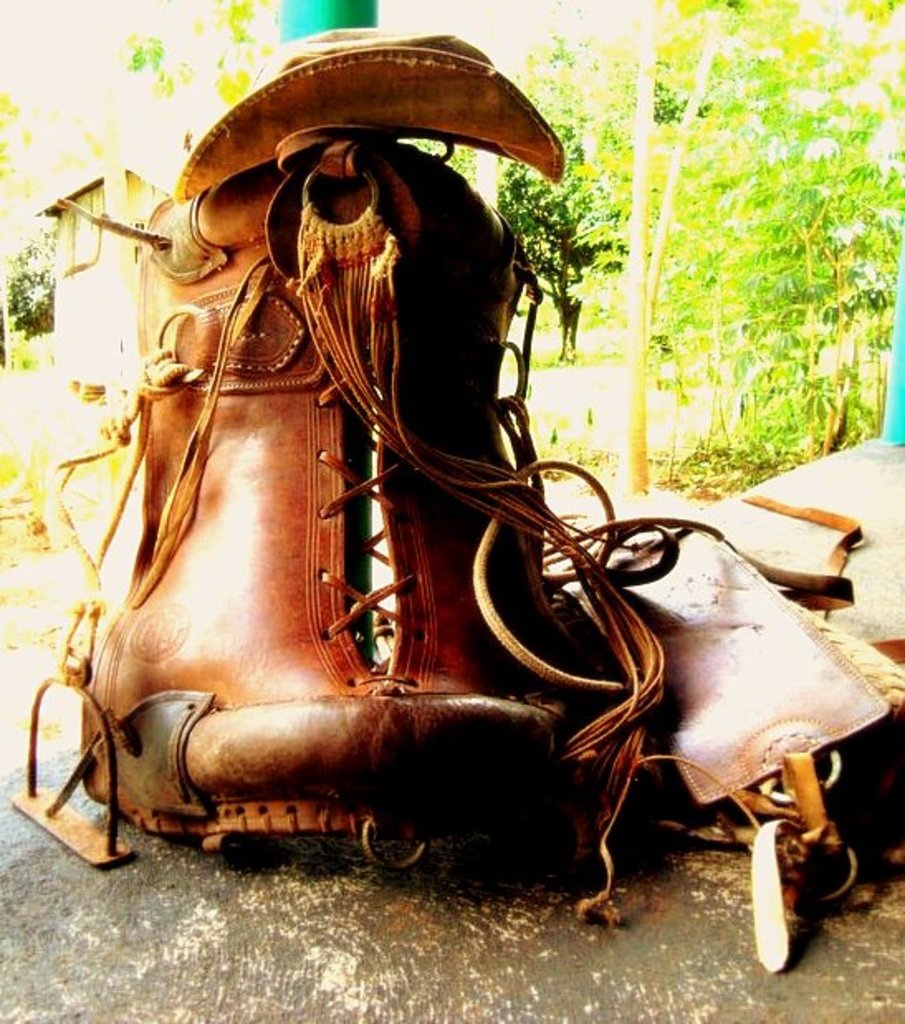}
    \end{subfigure}
    \hfill
    \begin{subfigure}[b]{0.23\textwidth}
        \centering
        \includegraphics[width=1\textwidth]{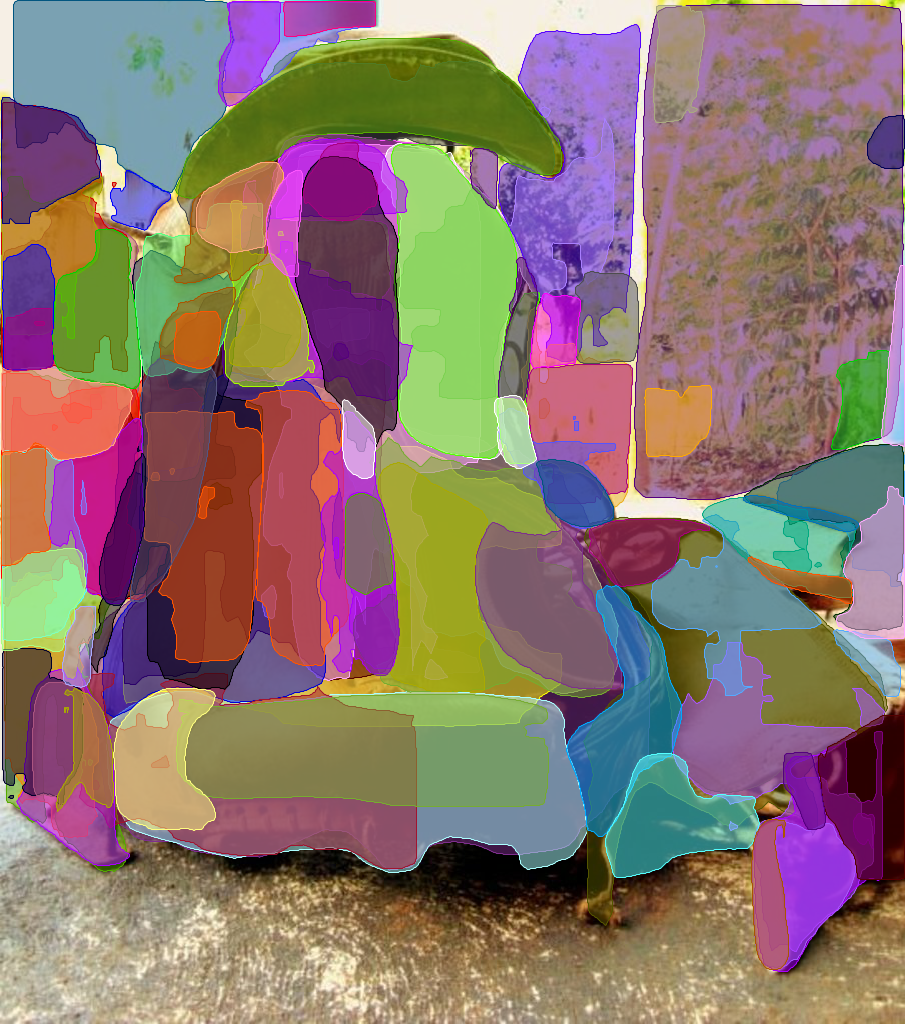}
    \end{subfigure}
    \hfill
    \begin{subfigure}[b]{0.23\textwidth}
        \centering
        \includegraphics[width=1\textwidth]{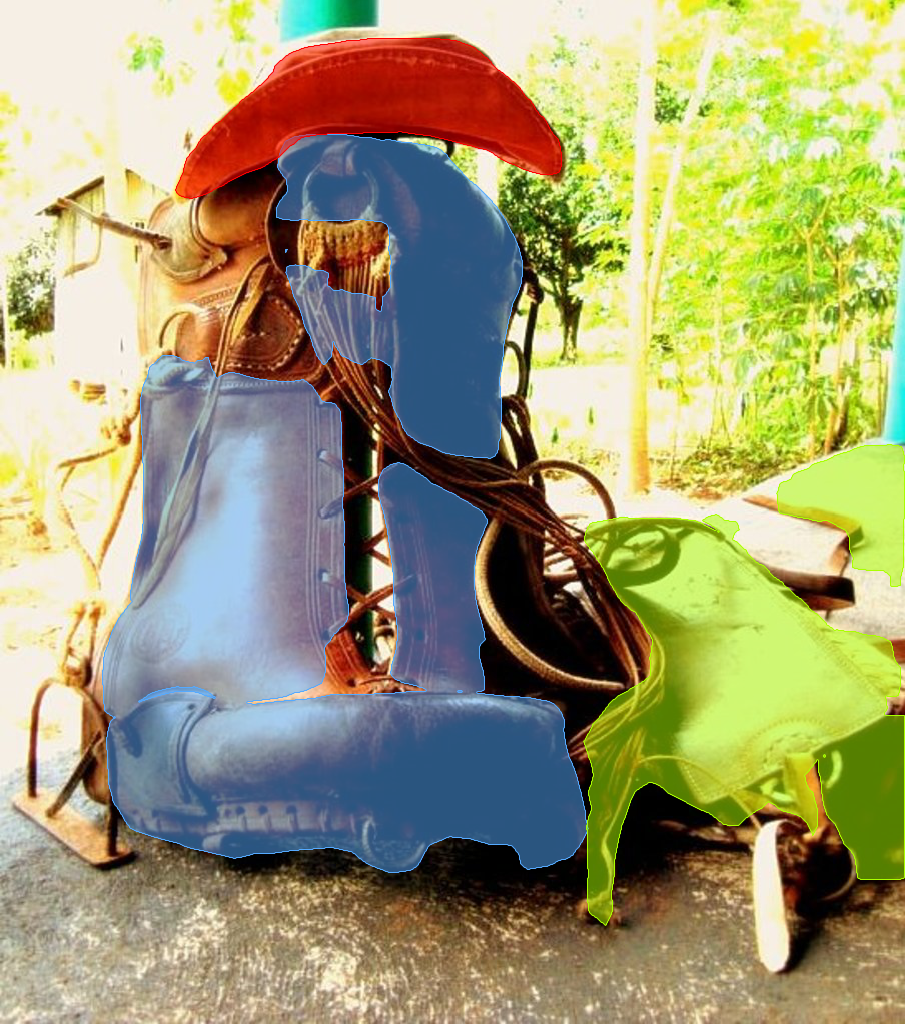}
    \end{subfigure}
    \hfill
    \begin{subfigure}[b]{0.23\textwidth}
        \centering
        \includegraphics[width=1\textwidth]{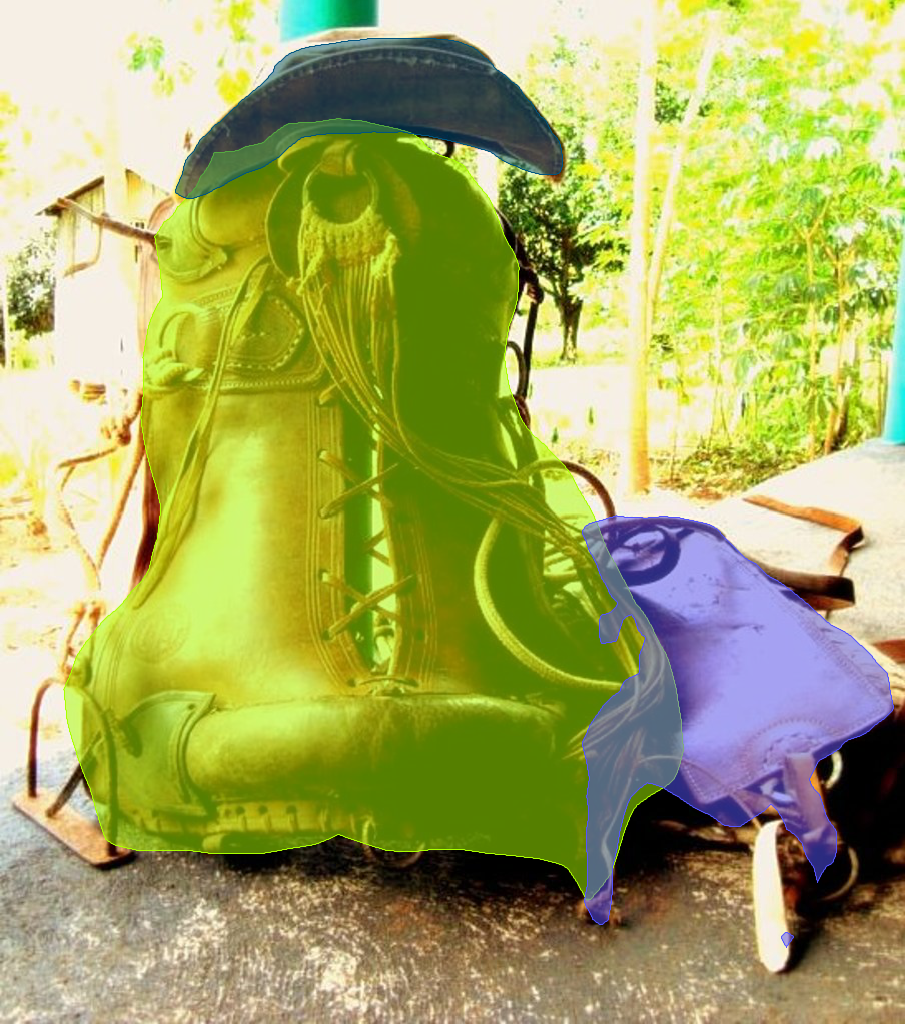}
    \end{subfigure}
    
    \vspace{4pt}
    \begin{subfigure}[b]{0.23\textwidth}
        \centering
        \includegraphics[width=1\textwidth]{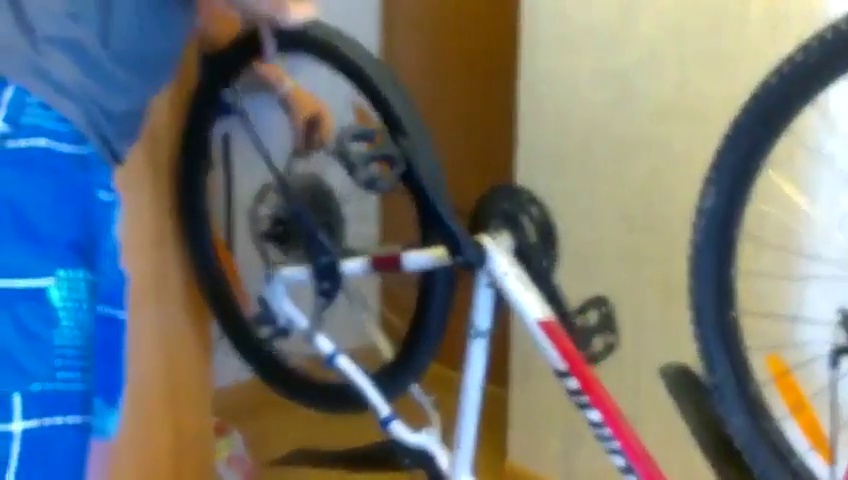}
    \end{subfigure}
    \hfill
    \begin{subfigure}[b]{0.23\textwidth}
        \centering
        \includegraphics[width=1\textwidth]{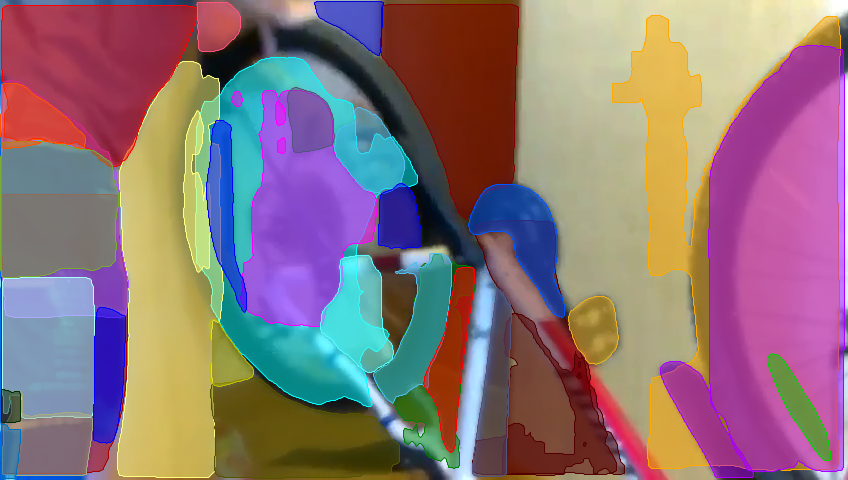}
    \end{subfigure}
    \hfill
    \begin{subfigure}[b]{0.23\textwidth}
        \centering
        \includegraphics[width=1\textwidth]{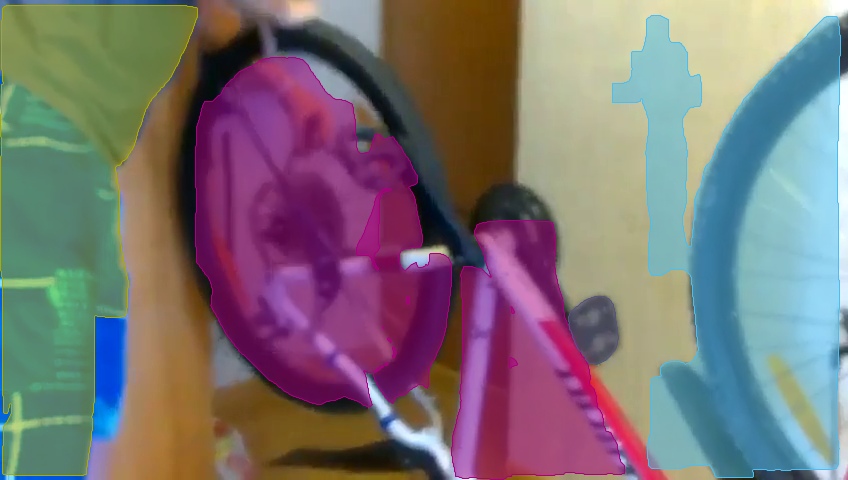}
    \end{subfigure}
    \hfill
    \begin{subfigure}[b]{0.23\textwidth}
        \centering
        \includegraphics[width=1\textwidth]{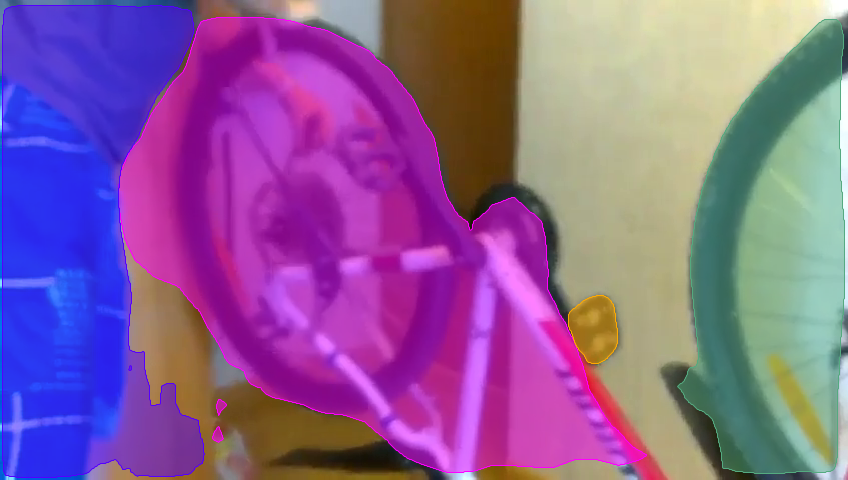}
    \end{subfigure}
    
    \vspace{4pt}
    \begin{subfigure}[b]{0.23\textwidth}
        \centering
        \includegraphics[width=1\textwidth]{figures/clustering/input_1335.jpg}
        \captionsetup{width=\textwidth, font=scriptsize}
        \subcaption{Input Image}
        \label{fig:inputimage}
    \end{subfigure}
    \hfill
    \begin{subfigure}[b]{0.23\textwidth}
        \centering
        \includegraphics[width=1\textwidth]{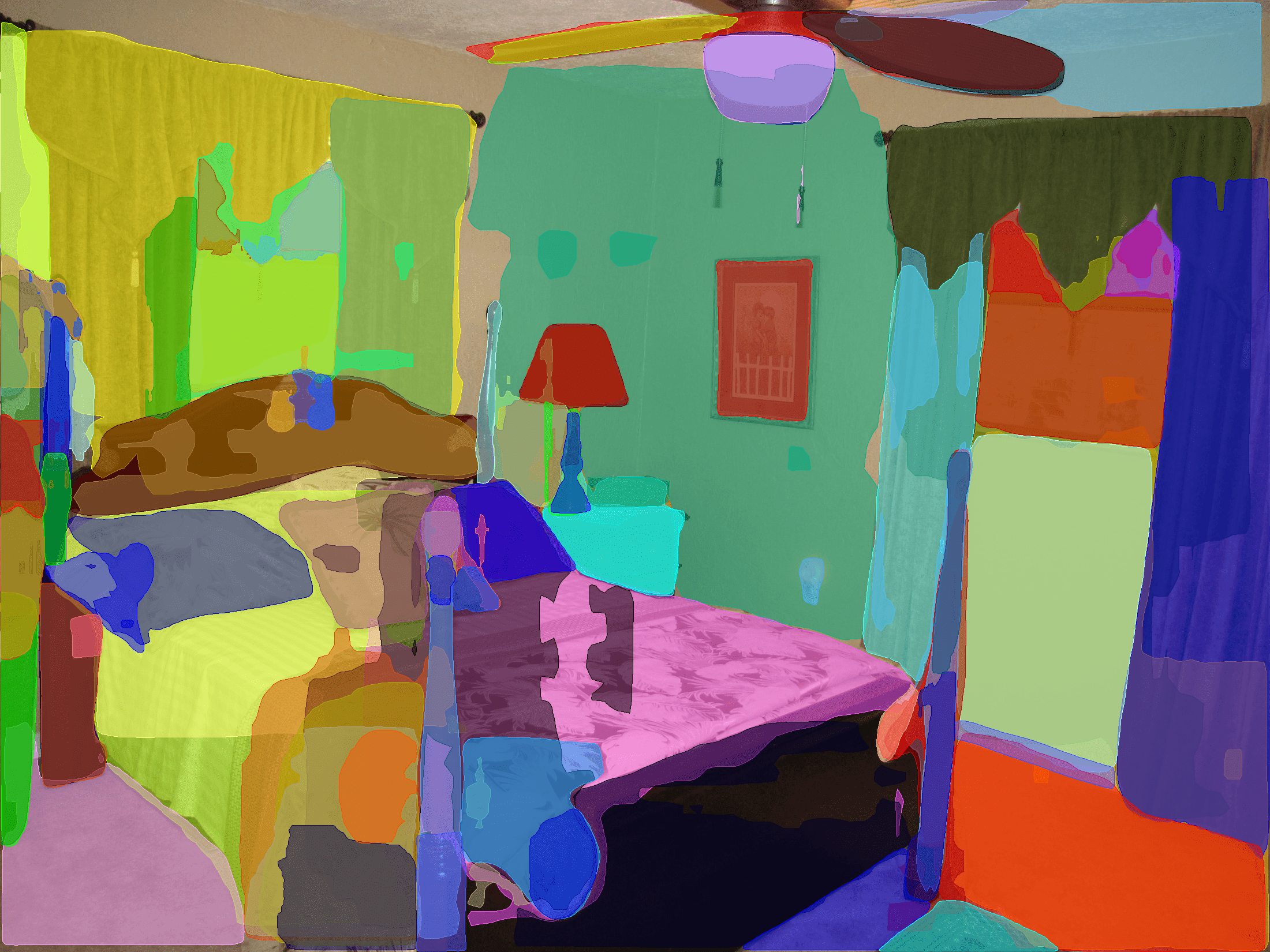}
        \captionsetup{width=\textwidth, font=scriptsize}
        \subcaption{part-mask output}
        \label{fig:partmasks}
    \end{subfigure}
    \hfill
    \begin{subfigure}[b]{0.23\textwidth}
        \centering
        \includegraphics[width=1\textwidth]{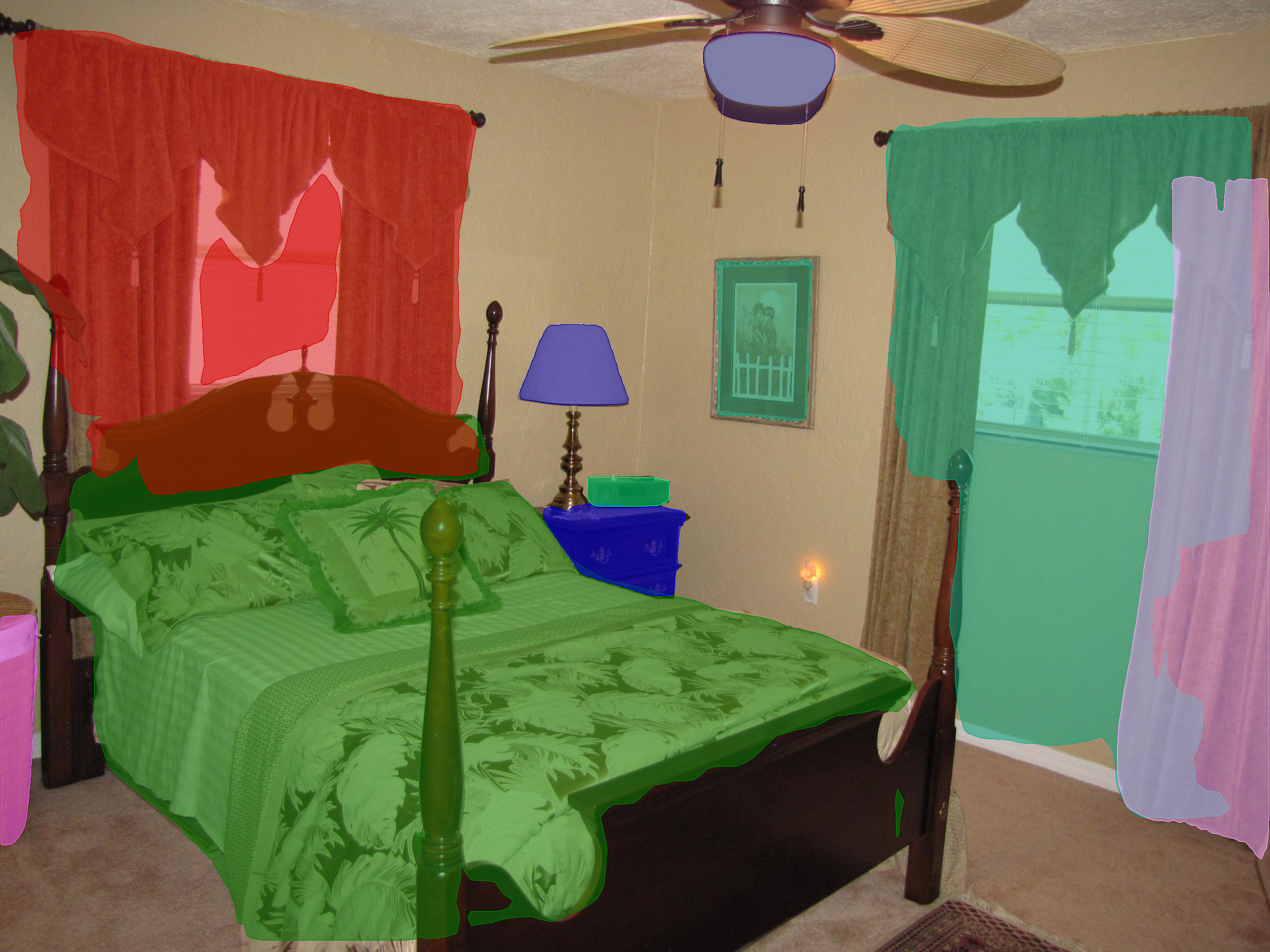}
        \captionsetup{width=\textwidth, font=scriptsize}
        \subcaption{Grouped prediction}
        \label{fig:mergedmasks}
    \end{subfigure}
    \hfill
    \begin{subfigure}[b]{0.23\textwidth}
        \centering
        \includegraphics[width=1\textwidth]{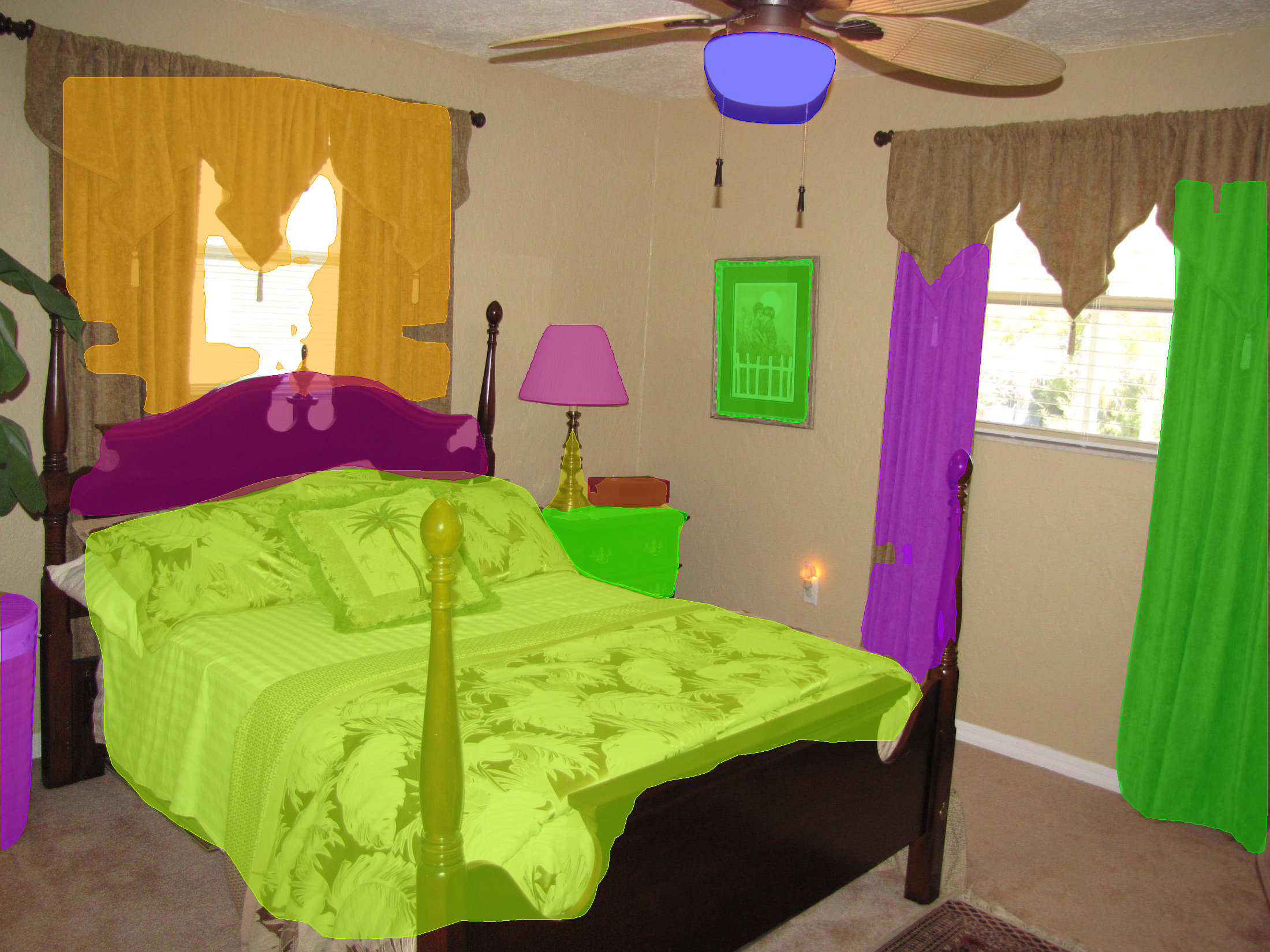}
        \captionsetup{width=\textwidth, font=scriptsize}
        \subcaption{Refined prediction}
        \label{fig:finaloutptus}
    \end{subfigure}
    
    \caption{{\bf Visualizing outputs after each stage of \Ours{} for cross-dataset training}. The images in the three rows belong to OpenImages~\cite{benenson2019large}, UVO~\cite{wang2021unidentified} and ADE20K~\cite{zhou2019semantic} datasets respectively. All outputs are generated by model trained on complete COCO dataset.
    (\subref{fig:inputimage}) shows the input image, followed by (\subref{fig:partmasks}) Part-mask prediction, (\subref{fig:mergedmasks}) grouped outputs using our affinity based grouping and (\subref{fig:finaloutptus}) refined prediction. The masks in last two columns correspond to true-positives with respect to the ground truth.}
    \label{fig:clusteringCrossDataset}
\end{figure*}

We visualize the outputs after each stage of UDOS for cross-category VOC to NonVOC setting in \figref{fig:cluster_voc} and for cross-dataset setting in \figref{fig:clusteringCrossDataset}. We illustrate the effect of our part-mask prediction module in in generating the segmentation masks for parts of objects, rather than the whole objects. This enables us to detect a much larger taxonomy of objects than what are present in the annotated concepts. For example, in \figref{fig:cluster_voc} for the case of cross-category transfer setting from VOC $\rightarrow$ Non-VOC, \textit{tie} and \textit{shoe} are not one of the annotated classes. Yet, our model effectively retrieves these from the image, instead of considering it a background or combining it with the boy. Our grouping module, powered by the context aggregation, then effectively groups the various part masks predicted on the \textit{pot}, \textit{boy} and \textit{tie}.
Note that the accuracy of predictions obtained by directly merging the part masks might be limited due to noisy part mask supervision, which are further corrected by our refinement layer. Similar observations for the cross-dataset setting are presented in \figref{fig:clusteringCrossDataset}.

\section{Qualitative comparisons}

\begin{figure*}[!t]
\begin{center}
        \centering
        \includegraphics[width=\textwidth]{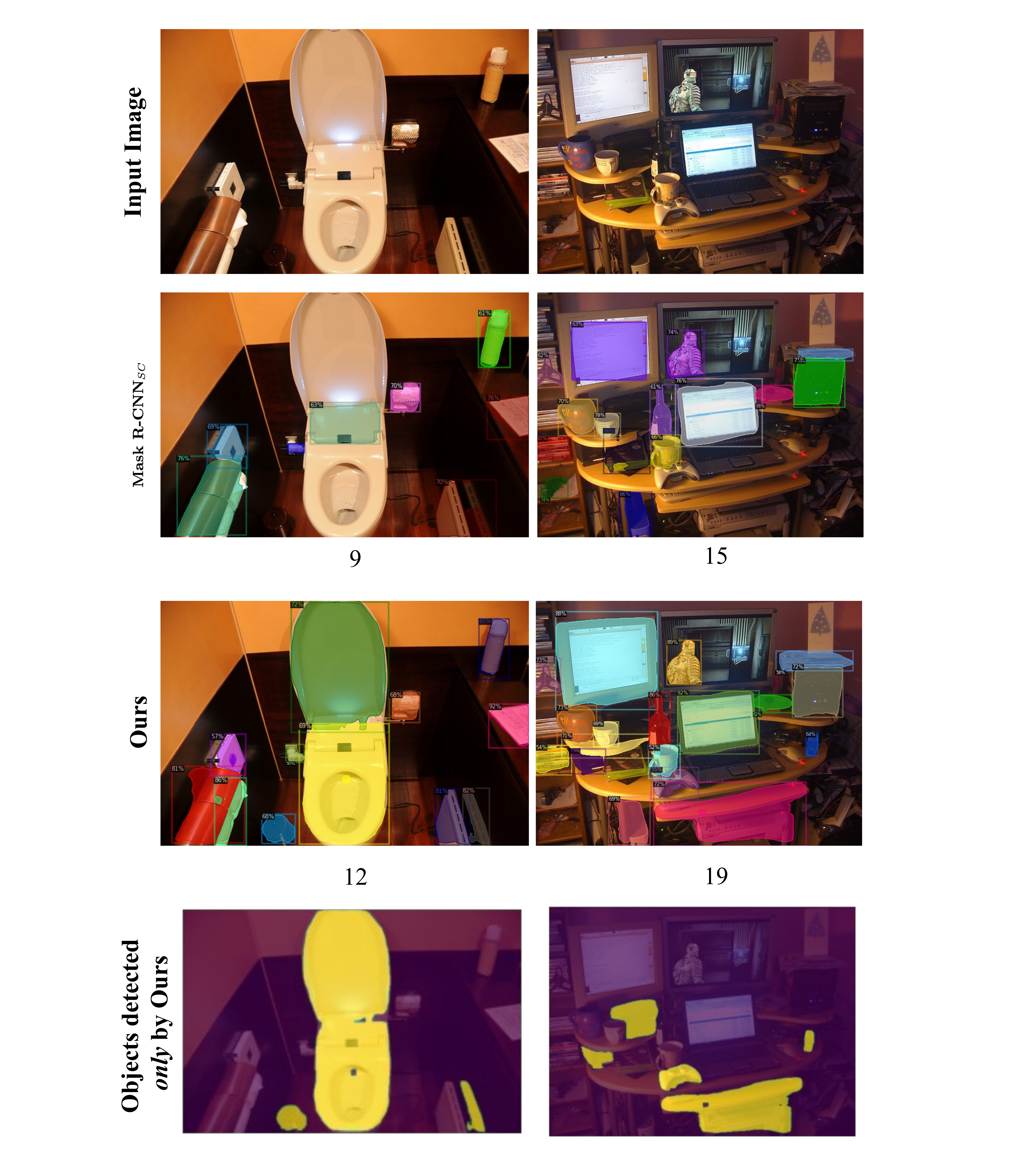}
        \captionsetup{width=\textwidth, font=footnotesize}
        \caption{{\bf Visualization of segmentations for model trained only on VOC classes} from COCO dataset. For various input images given in the first row, the second row shows result using Mask-RCNN$_{SC}$, third row shows output using \Ours{} and the fourth row shows some predictions made only by \Ours{} and missed by Mask-RCNN$_{SC}$ on these images. We also show the number of detections made by the network below each image. All images belong to COCO dataset.
        }
        \label{fig:visualization_supp_coco}
\end{center}
\end{figure*}

\begin{figure*}[!t]
\begin{center}
        \centering
        \includegraphics[width=0.8\textwidth]{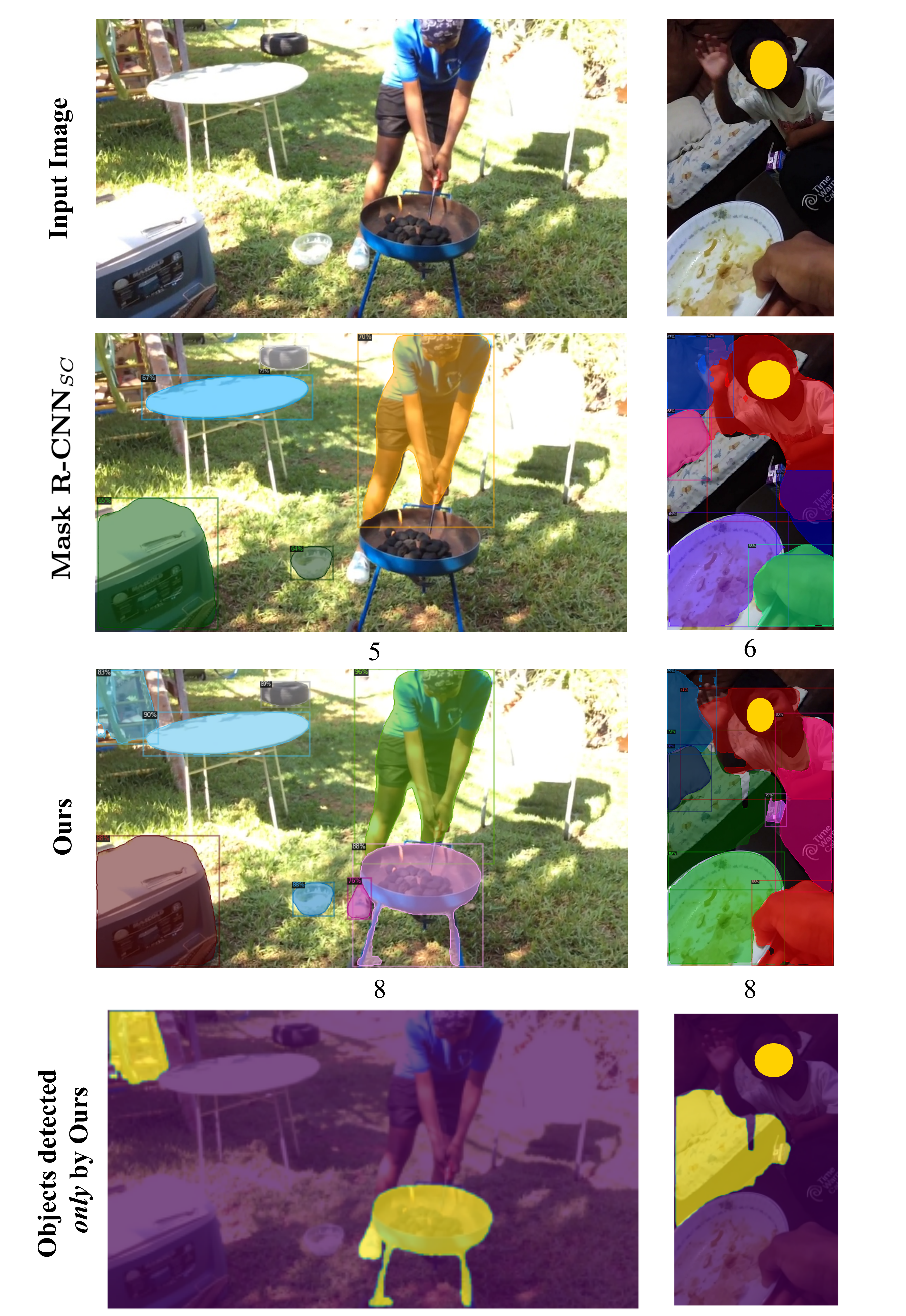}
        \captionsetup{width=\textwidth, font=footnotesize}
        \caption{{\bf Visualization of segmentations for model trained on all COCO classes}. For various input images given in the first row, the second row shows result using Mask-RCNN$_{SC}$, third row shows output using \Ours{} and the fourth row shows some predictions made only by \Ours{} and missed by Mask-RCNN$_{SC}$ on these images. We also show the number of detections made by the network below each image. All images belong to UVO dataset.
        }
        \label{fig:visualization_supp_uvo}
\end{center}
\end{figure*}

In addition to the comparisons provided in the main paper, we provide more comparisons of predictions made by \Ours{} and Mask R-CNN$_{SC}$ in \figref{fig:visualization_supp_coco} for the setting where we train only using VOC categories from COCO. We also show the predictions made on the cross-dataset setting, by using a model trained on all COCO categories and testing on images from UVO~\cite{wang2021unidentified} in \figref{fig:visualization_supp_uvo}. 
In each case, we also show the predictions made \textit{only} by \Ours{} and missed by Mask R-CNN$_{SC}$ (highlighted in yellow), indicating the utility of our approach on open world instance segmentation.

For instance, in the second column in \figref{fig:visualization_supp_coco}, the predictions made by Mask R-CNN$_{SC}$ do not include objects like \textit{keyboard}, \textit{joystick}, \textit{glass} and \textit{speaker} which are efficiently retrieved by \Ours{}. Also note that the number of predictions made by \Ours{} is always higher than Mask R-CNN$_{SC}$ for both cases of cross-category transfer in \figref{fig:visualization_supp_coco} as well as cross-dataset transfer in \figref{fig:visualization_supp_uvo}.

\end{document}